%% file: main.tex
\documentclass[10pt,twocolumn,letterpaper]{article}
\usepackage{cvpr}              % To produce the CAMERA-READY version
\input{sec/_1_preamble}

\definecolor{cvprblue}{rgb}{0.21,0.49,0.74}
\usepackage[pagebackref,breaklinks,colorlinks,allcolors=cvprblue]{hyperref}
\usepackage{booktabs}

\def\paperID{2264} % *** Enter the Paper ID here
\def\confName{CVPR}
\def\confYear{2025}

\title{\includegraphics[height=5mm]{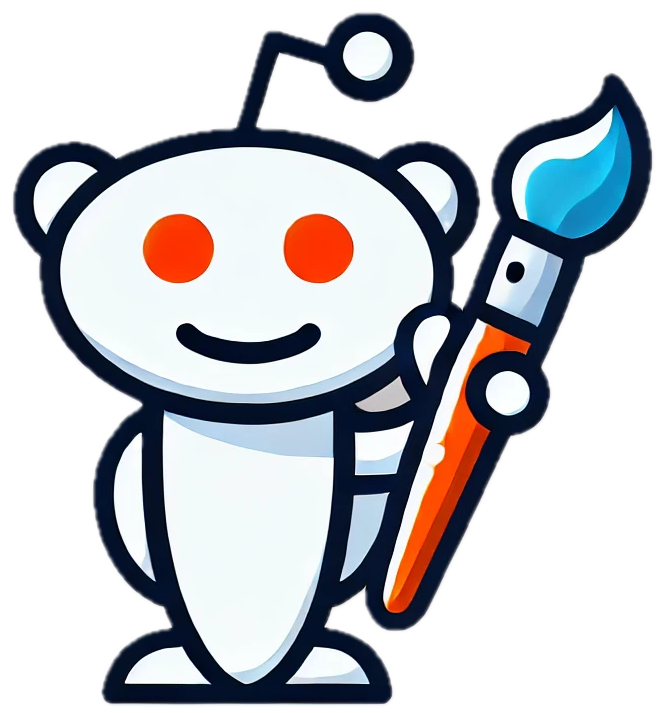} 
\textsc{RealEdit}: Reddit Edits As a Large-scale\\
Empirical Dataset for Image Transformations}

\author{%
  {\bf Peter Sushko}$^{\dagger}$, {\bf Ayana Bharadwaj}$^{\dagger}$, {\bf Zhi Yang Lim}$^{\dagger}$, {\bf Vasily Ilin}$^{\dagger}$, {\bf Ben Caffee}$^{\dagger}$,\\ 
  {\bf Dongping Chen}$^{\dagger}$, {\bf Mohammadreza Salehi}$^{\dagger}$, {\bf Cheng-Yu Hsieh}$^{\dagger}$, {\bf Ranjay Krishna}$^{\dagger,\psi}$\\[1ex]
  $^{\dagger}$University of Washington \quad $^{\psi}$Allen Institute for AI
}

\newcommand{\ours}{\textsc{RealEdit}\xspace}

\begin{document}
\pagestyle{plain}
\twocolumn[{%
\renewcommand\twocolumn[1][]{#1}%
\maketitle
\includegraphics[width=1\linewidth]{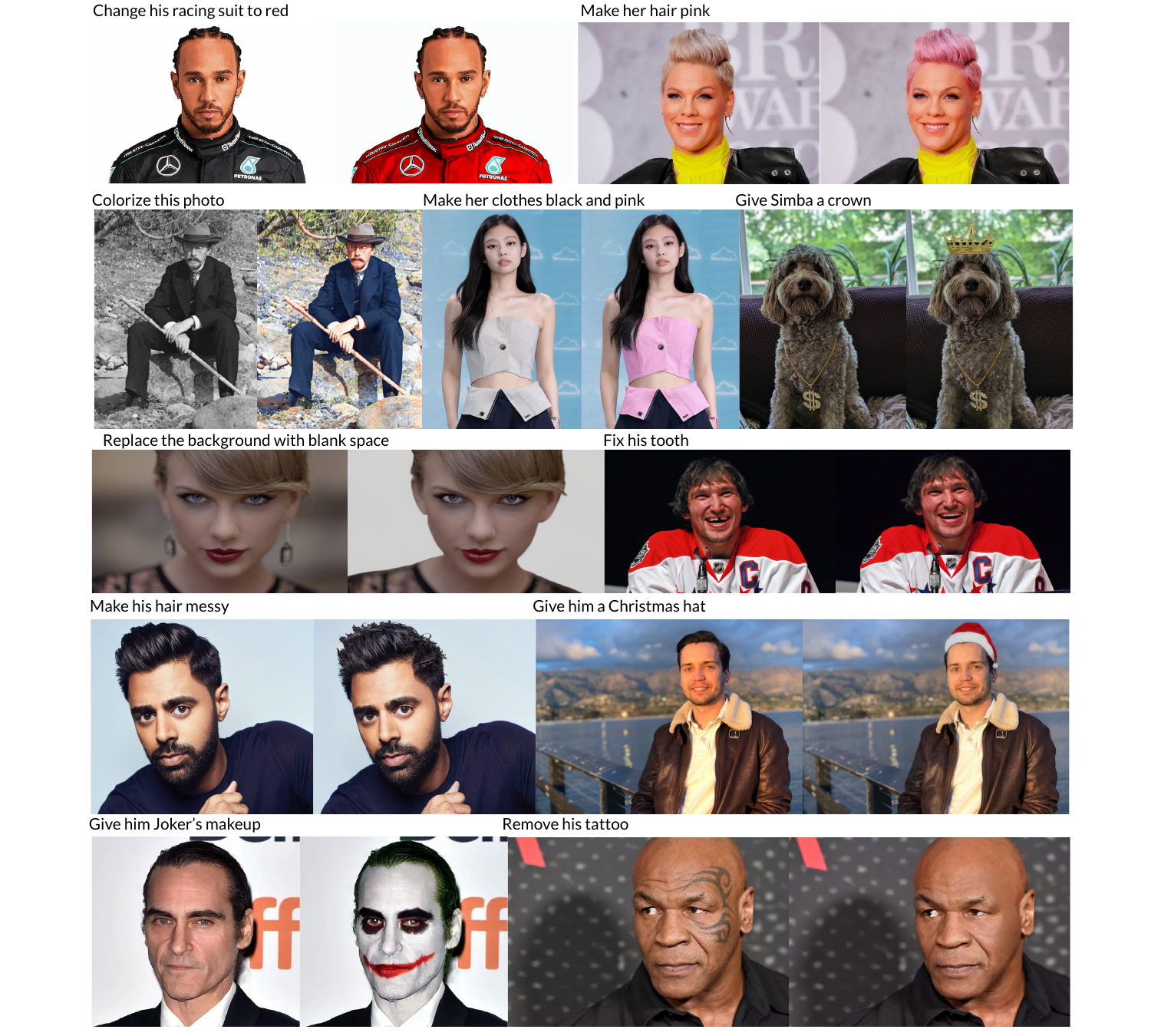}
\captionof{figure}{\textbf{We visualize edits made by our model.} We introduce \RealEdit, a large-scale image editing dataset sourced from Reddit with real-world user edit requests and human-edits. By finetuning on \RealEdit, our resultant model outperforms existing models by up to 165 Elo points with human judgment and delivers real world utility to real user requests online.}
\label{fig:teaser}
}]

\begin{abstract}
    \input{sec/00_abstract}
\end{abstract}

\input{sec/01_intro.tex} 
\input{sec/02_related_works.tex}

\input{sec/03_dataset.tex}

\input{sec/04_analysis.tex}

\input{sec/05_method.tex}

\input{sec/06_experiments.tex}
\input{sec/07_discussion.tex}

\clearpage
{
    \small
    \bibliographystyle{ieeenat_fullname}
    \bibliography{main}
}

\input{sec/X_suppl}
\end{document}

%% file: sec/_1_preamble.tex
\usepackage{multirow}
\usepackage{pifont}
\usepackage{tabularray}
\usepackage{graphicx} % For including images
\usepackage{float}    % For controlling float placement
\usepackage{afterpage}
\usepackage{xspace}
\usepackage{xcolor}
\definecolor{darkred}{rgb}{0.7, 0.0, 0.0}
\definecolor{darkgreen}{rgb}{0.0, 0.5, 0.0}
\usepackage{multicol}
\usepackage{afterpage}
\usepackage{caption}
\usepackage{rotating}

\newcommand{\RealEdit}{\textsc{RealEdit}\xspace}
\newcommand{\truemedia}{TrueMedia.org\xspace}

%% file: sec/00_abstract.tex
Existing image editing models struggle to meet real-world demands; despite excelling in academic benchmarks, we are yet to see them adopted to solve real user needs. 
The datasets that power these models use artificial edits, lacking the scale and ecological validity necessary to address the true diversity of user requests. 
In response, we introduce \ours, a large-scale image editing dataset with authentic user requests and human-made edits sourced from Reddit. \ours contains a test set of $9.3$K examples the community can use to evaluate models on real user requests. 
Our results show that existing models fall short on these tasks, implying a need for realistic training data.
So, we introduce $48$K training examples, with which we train our \ours model. 
Our model achieves substantial gains—outperforming competitors by up to $165$ Elo points in human judgment and $92\%$ relative improvement on the automated VIEScore metric on our test set. We deploy our model back on Reddit, testing it on new requests, and receive positive feedback. 
Beyond image editing, we explore \ours’s potential in detecting edited images by partnering with a deepfake detection non-profit. Finetuning their model on \ours data improves its F1-score by $14$ percentage points, underscoring the dataset's value for broad, impactful applications.~\footnote{\scriptsize Project page: \url{https://peter-sushko.github.io/RealEdit}}

%% file: sec/01_intro.tex
\section{Introduction}

The need to edit photos is more important than ever—people everywhere seek to perfect, enhance, or restore their images, from casual snapshots to treasured memories. 
If more effective and aligned editing models were readily available, many would use them for a variety of purposes: to remove an unwanted photobomber, adjust lighting in their selfies, restore their grandparents' wedding photos, or even add creative effects.
% Text-guided image editing has gained remarkable popularity by offering users an intuitive way to specify these transformations: through natural language prompts~\cite{brooks2023instructpix2pix, zhang2024magicbrush}.
This demand is vividly demonstrated in online communities like Reddit’s \textit{r/PhotoshopRequest}\footnote{\scriptsize 
 \url{https://www.reddit.com/r/PhotoshopRequest}} and \textit{r/estoration}\footnote{\scriptsize 
 \url{https://www.reddit.com/r/estoration}}, with over 1.5 million combined members. Many users pay money for quality edits, highlighting the demand for advanced, user-friendly editing tools.

\begin{figure*}[!h]
    \centering
    \includegraphics[width=\textwidth]{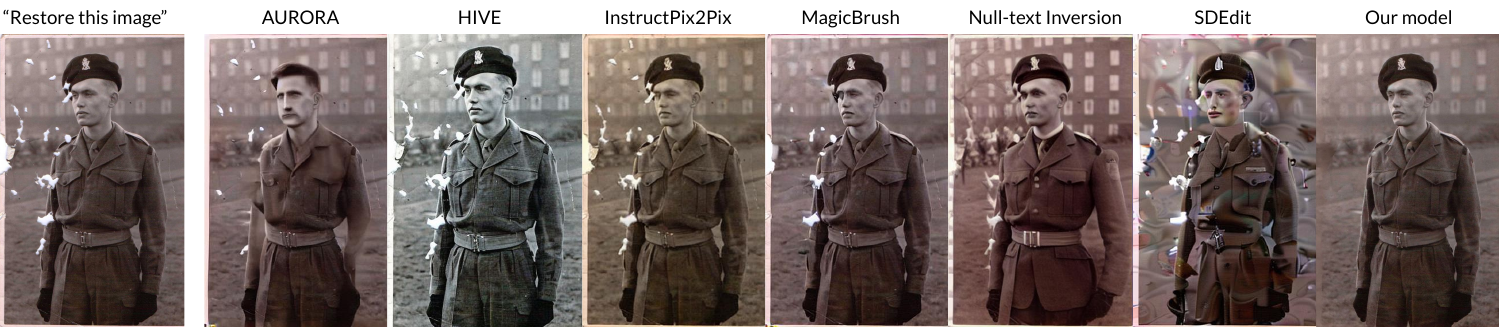}
    \caption{Baselines struggle on simple, practical tasks, such as restoring a damaged photograph. Our model is successful.}
    \label{fig:soldier}
\end{figure*}

Despite the impressive capabilities in image generation and modification led by recent advancement of diffusion models~\cite{Ramesh2022DALLE2, rombach2022high, brooks2023instructpix2pix, zhang2024magicbrush}, seemingly straightforward real-world editing tasks, like ones from the Reddit's \textit{r/PhotoshopRequest}, continue to pose significant challenges to existing models.
For instance, while existing models are effective at artistic transformations or generating stylized content~\cite{ zhang2024magicbrush, mokady2023null, krojer2024learning, meng2021sdedit, sheynin2024emu}, they fall short at some of the most common real-world requests such as restoring a damaged image (see Figure~\ref{fig:soldier}). 
This discrepancy highlights a critical misalignment between the capabilities of current editing models and the actual needs of users.

One major challenge for models to effectively tackle real-world image editing is the diversity and open-ended nature of the tasks involved. However, most existing models are trained with synthetic or arbitrarily created datasets that do not characterize human-centered objectives well, as is shown in Table \ref{tab:dataset_comparison}.
For example, in Ultra-Edit~\cite{zhao2024ultraedit}, ``adding a rainbow'' to an image constitutes a significant portion of the data set. As a result, models trained on these datasets struggle to address the practical needs of real-world users.

In this work, we introduce \ours, a large-scale text-guided image editing dataset meticulously compiled from Reddit. \ours, by design, more faithfully reflects the distribution of image editing needs.
Specifically, we source image editing requests from two of the largest relevant subreddit communities, \textit{r/PhotoshopRequest} and \textit{r/estoration}, into a dataset consisting of over $57$K editing examples, wherein each example comprises of an input image, an instruction, and one or multiple edits performed by humans. Overall, there are a total of $151$K input and edited images in this collection.
By carefully preprocessing and filtering out ambiguous and noisy examples with meticulous manual verification, we transform part of the collected examples in \ours into an evaluation set that consists of more than $9.3$K real-world image editing requests to test models' capability.
Notably, \ours evaluation set shows that real-world requests differ drastically from existing evaluation datasets~\cite{zhang2024magicbrush, sheynin2024emu}, on which existing models struggle.

To build an effective image editing model for real-world tasks, we finetune a new text-guided image editing model, on \ours's training examples. To produce useful edits that preserve the identities of the people in photos, we upgrade InstructPix2Pix~\cite{brooks2023instructpix2pix} by replacing its Stable Diffusion~\cite{rombach2022high} decoder with OpenAI's Consistency decoder~\cite{openai_consistencydecoder}, which was pretrained on more human-centric data.

Our model demonstrates significantly better performance than existing state-of-the-art models on \ours's test set with a human preference (N=4,196) Elo score of $1184$, beating the next best model by $165$ points. We also outperform existing models using automated metrics: our model achieves $4.61$ VIEScore versus the next best score of $2.4$ (amongst other metrics).
Moreover, our model still remains competitive with MagicBrush and EmuEdit~\cite{zhang2024magicbrush, sheynin2024emu} on their test sets. We further validate our model by completing new Reddit requests and receiving positive feedback.

Finally, we partner with \truemedia, a non-profit aimed at AI-generated content detection. By adding human-made edits from our dataset, we improve their model's F1-score by $14$ points. Our ecologically valid experimentation highlights the dataset's value outside of editing tasks.

%% file: sec/02_related_works.tex
\section{Related work}
\noindent\textbf{Image editing datasets.}
While extensive datasets exist for captioning and identifying edited images within fixed domains~\cite{desai2021redcaps, park2018double}, there is a notable lack of large-scale, human-edited image datasets.
% Creating these datasets is challenging due to the labor-intensive nature of manual editing. 
% For model training, high-quality image editing datasets are essential.
Currently, larger-scale image editing datasets mostly rely on synthetic data~\cite{brooks2023instructpix2pix, zhang2024magicbrush, sheynin2024emu, zhang2024hive, zhao2024ultraedit}, while the ones with human edited images are limited in size~\cite{shi2020benchmark, tan2019expressing}.
While synthetic datasets may include human inputs, such as generating instructions or ranking edits \cite{zhang2024magicbrush, zhang2024hive, brooks2023instructpix2pix}, these datasets do not contain \textit{edits} that are completed by humans.
Most importantly, existing datasets are curated in ways that do not necessarily characterize real-world editing distribution well. We compare \RealEdit to existing datasets in Table~\ref{tab:dataset_comparison}.
% Table \ref{tab:dataset_comparison} compares existing image editing datasets, illustrating that there is no large, fully human-edited dataset with ground-truth edits, a gap that restricts the domain of current editing models to arbitrary cases and inferior quality.

% As a result, edited images may contain varying quality model performance suffers.
% Most importantly, while notable datasets like GIER \cite{shi2020benchmark} and the Image Editing Request Dataset \cite{tan2019expressing} contain around 6,000 and 4,000 samples respectively, they are insufficient for training robust models that capture the wide range of real-world editing use cases.

\paragraph{Text-guided image editing.}
There is a rich literature in models focusing on specific image editing tasks, such as inpainting~\cite{yu2018generative}, denoising~\cite{goyal2020image}, and style transfer~\cite{gatys2016image}.
Recent advancements emphasize generalized models that better align with human use cases, leading to innovative methods such as generating programs to modify images~\cite{gupta2023visual}, as well as end-to-end diffusion-based or GAN-based editing models~\cite{karras2019style, patashnik2021styleclip, avrahami2022blended, meng2021sdedit, wang2023imagen, xie2023smartbrush}. Diffusion models like Stable Diffusion~\cite{rombach2022high} excel at generating images from text prompts,
serving as versatile models for image generation~\cite{zhan2023multimodal}.
Several models~\cite{brooks2023instructpix2pix, mokady2023null, kawar2023imagic} utilize diffusion-based techniques for editing, though generating images from captions alone may compromise fidelity. To mitigate this, some models~\cite{brooks2023instructpix2pix, mokady2023null, zhang2024magicbrush, krojer2024learning} leverage Prompt-to-Prompt technique~\cite{hertz2022prompt}, employing cross-attention maps to preserve most of the original image. Others achieve consistency by fine-tuning diffusion models to reconstruct images using optimized text embeddings, blending these with target text embeddings~\cite{kawar2023imagic}.
However, limitations persist, such as stuggles with face generation~\cite{borji2022generated} and cross-attention requiring minimal, often single-token caption variation.

\paragraph{Evaluating image editing models.}
% \textbf{VQA-based Evaluation.}
Originating from early text summarization in NLP~\citep{Narayan2018RankingSF}, QA-based evaluation methods automatically transform prompts into questions and use them to validate generated content~\citep{Durmus2020FEQAAQ, Deutsch2020TowardsQA, Eyal2019QuestionAA}. 
In text-to-image generation, VQA-based evaluation methods transfer text into atomic questions and conduct VQA to verify generated images, providing enhanced fine-grained and interpretable benchmark results~\citep{cho2023davidsonian, lin2024evaluating, chen2024mllm}.
Notably, TIFA~\citep{hu2023tifa} pioneered the use of VQA for automatic evaluation, while subsequent works enhanced model-human correlation~\citep{yarom2024you, lu2024llmscore}, incorporated additional modules and MLLM-as-a-Judge~\citep{ghosh2024geneval, cho2024visual, ye2024justice, chen2024mllm, ku2023viescore}.
To evaluate image editing models, we follow and extend existing work~\cite{sheynin2024emu} in casting the evaluation into image generation evaluation wherein we measure the faithfulness of the edited images to their target output captions, using the aforementioned VQA-based frameworks.

% \textbf{VIEScore}

%% file: sec/03_dataset.tex
\section{\RealEdit}
\label{sec:realedit}

We introduce \RealEdit: a high-quality large-scale dataset for text-guided image editing. \RealEdit dataset includes 48K training data points and 9K test data points, each featuring an \textit{original image}, an \textit{editing instruction}, and one to five \textit{human-edited output images}. Altogether, we are publishing a total of 151K images. \RealEdit is the first large-scale image editing dataset wherein real-world users both submit and complete the requests (Table \ref{tab:dataset_comparison}).

\input{tables/image_editing_datasets}

\begin{figure*}[!h]
    \centering
    \includegraphics[width=\textwidth]{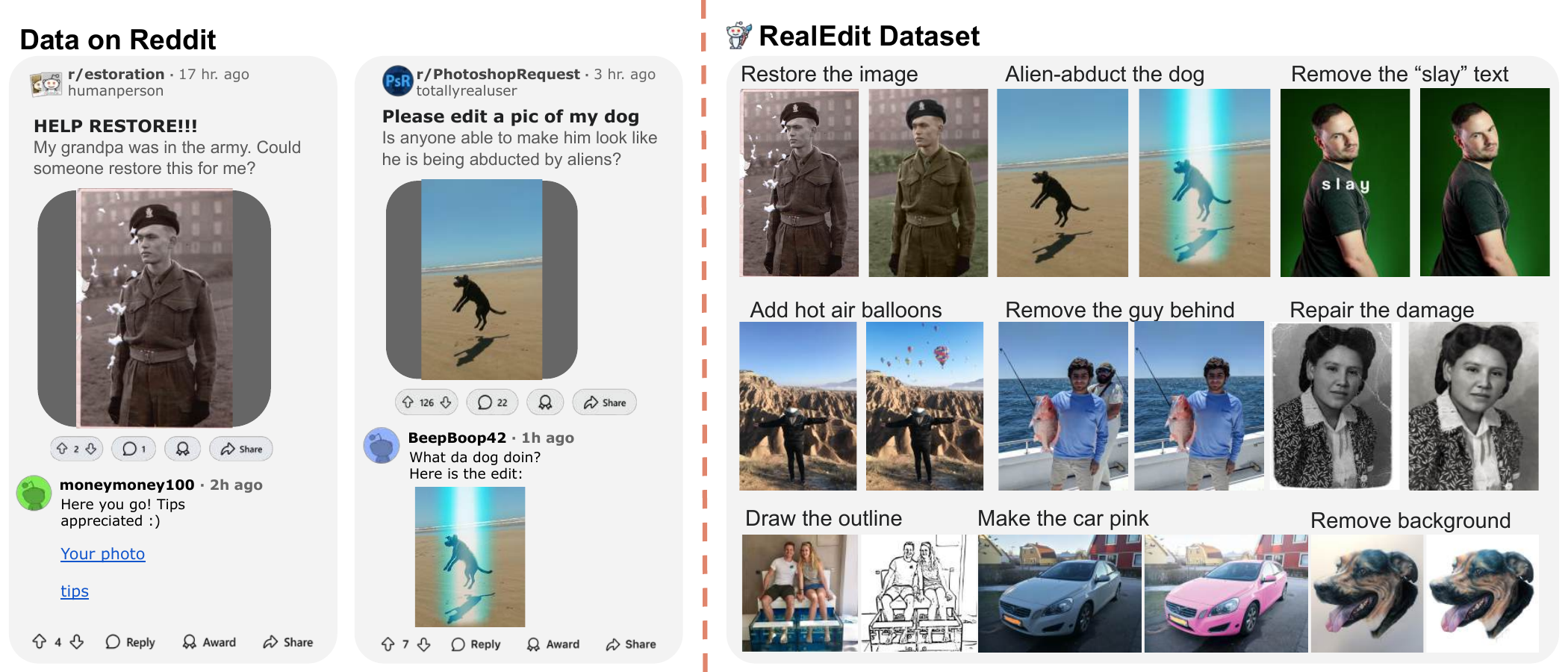}
    \caption{\textbf{Dataset curation pipeline.} We source data from r/estoration and r/PhotoshopRequest. From the posts, we extract input images and edit instructions. The instructions are processed using a VLM to isolate the editing task. From the comments, we collect up to 5 human-edited outputs per post.}
    \label{fig:data_pipeline}
    \vspace{-2mm}
\end{figure*}

\subsection{Dataset creation pipeline}
The extensive and structured nature of Reddit makes it an ideal source for creating diverse large-scale datasets rooted in real-world content. We leverage this by developing a data collection pipeline with three key steps: (1) collecting raw post and comment data from the subreddits of interest, (2) processing and organizing the data, and (3) manual verification to ensure safe and high-quality outputs (Figure~\ref{fig:data_pipeline}).
\noindent\textbf{Step 1: Subreddit selection.}
We build a diverse image editing dataset from two key subreddits to cover a wide range of tasks. The main source, \href{https://www.reddit.com/r/PhotoshopRequest/}{r/PhotoshopRequest}, provides 261K posts and 1.1M comments on tasks ranging from object removal and background changes, to creative edits.
Additionally, we source requests from \href{https://www.reddit.com/r/estoration/}{r/estoration} for their sentimental value to users.
This subreddit contributes 20K posts and 126K comments focused on restoring old photos, including repairing creases, colorizing black-and-white images, and enhancing clarity.
We exclude larger communities like \href{https://www.reddit.com/r/photoshopbattles/}{r/photoshopbattles} due to their emphasis on humor and less specific editing needs. The dataset consists of original image URLs sourced from posts, edit instructions, and edited image URLs taken from the comments. The images we collected were posted between 2012 and 2021 which implies low likelihood of AI-generated content.

\noindent\textbf{Step 2: Instruction refinement and caption generation.}
One challenge in collecting web-crawled data is that user-provided instructions may be noisy, often including personal anecdotes or task-irrelevant details
(\textit{e.g., ``This photo was taken of my Mother and me at my Grandmother’s wake. I would love to get this framed for my Mom’s birthday next month. I love the photo, but the person who took it put filters all over it. I was wondering if someone could make it look more natural.''}). We use GPT-4o~\cite{openai2023gpt4} to summarize the text to only the key editing requirements. The pipeline refines the noisy instruction above into the following: ``Restore image damage and enhance clarity''.

For the \RealEdit test set, we generated captions for both input and edited images using vision-language models to support evaluation on caption-based metrics. Implementation details are provided in Appendix B.
% \ref{supp:data_preprocess}.

% For building the \RealEdit test set, we additionally add captions for both the input and the edited images, in order to facilitate evaluation metrics that operate on the input or target output captions. 
% % specifically for the test split in order to facilitate evaluation for models that require captions.
% The input image caption is generated using LLaVA-Next~\cite{liu2024llavanext}, conditioned on the processed instruction. This instruction-aware captioning is particularly useful for subtle edits, as it ensures the LLaVA-Next model to focus on specific areas of the image that might otherwise be overlooked.
% For the edited images, we combined the input caption with the editing instructions and used GPT-4o~\cite{openai2023gpt4} to generate a caption that summarizes both the original content and the modifications made. 

% \begin{figure*}[htbp]
%     \centering
%     \includegraphics[width=\textwidth]{figs/instruction-aware-vlm.png}
%     \caption{Conditioning the VLM on the editing instruction improves caption quality.}
%     \label{fig:instruction-aware-vlm}
% \end{figure*}

% \begin{figure*}[htbp]
%     \centering
%     \includegraphics[width=\textwidth]{figs/output_caption.png}
%     \caption{Captions for edited images are produced by feeding the edit instruction and original image caption to GPT-4o.}
%     \label{fig:output_caption}
% \end{figure*}

\noindent\textbf{Step 3: Data verification and final composition.}
After generating the dataset, we conducted a rigorous multi-stage verification process to ensure data quality. All images were screened for inappropriate content using the opennsfw2~\cite{bhky_opennsfw2} network, filtering out those flagged as explicit. Additionally, the \RealEdit test set was manually reviewed by two annotators evaluating the following criteria:
(1) appropriateness of the input image, (2) applicability of the instruction, and (3) correctness of the output image. We employed qualitative coding, establishing four distinct codes, with Cohen’s Kappa yielding 0.61, indicating substantial agreement. Approximately 78\% of the data points were agreed upon as high quality and included in the final test set for \RealEdit, mitigating individual biases.

% Through the manual review process, approximately 78\% of the images were agreed upon by two independent annotators as high-quality and were included in the final test set for \RealEdit.

% \begin{figure*}[htbp]
%     \centering
%     \includegraphics[width=\textwidth]{figs/template_data_pipeline.pdf}
%     \caption{Template of data collection pipeline}
%     \label{fig:template_data_pipeline}
% \end{figure*}

%% file: tables/image_editing_datasets.tex
\begin{table}[]
    \vspace{-4mm}
    \centering
    \caption{Size and human involvement across editing datasets. \RealEdit is the largest dataset containing human-made edits.}  
    \label{tab:dataset_comparison}
    
    \resizebox{\linewidth}{!}{%
    \begin{tabular}{lr cccc}
    \toprule
                   \multirow{2}{*}{Dataset} & \multirow{2}{*}{Size}   & Large     & Human    & Human & Real-world  \\ 
                    & & scale? & select outputs? & edit? & requests? \\
    \midrule
    InstructPix2Pix~\cite{brooks2023instructpix2pix} & 454K     & \textcolor{darkgreen}{\ding{51}}    & \textcolor{darkred}{\ding{55}}          & \textcolor{darkred}{\ding{55}}& \textcolor{darkred}{\ding{55}}\\ 
    MagicBrush~\cite{zhang2024magicbrush}     & 10K      &  \textcolor{darkred}{\ding{55}}  & \textcolor{darkred}{\ding{55}}         & \textcolor{darkred}{\ding{55}} & \textcolor{darkred}{\ding{55}}\\ 
    EmuEdit~\cite{sheynin2024emu}         & 10M          & \textcolor{darkgreen}{\ding{51}} & \textcolor{darkred}{\ding{55}} & \textcolor{darkred}{\ding{55}} & \textcolor{darkred}{\ding{55}}\\ 
    HIVE~\cite{zhang2024hive}            & 1.1M         & \textcolor{darkgreen}{\ding{51}} & \textcolor{darkgreen}{\ding{51}}         & \textcolor{darkred}{\ding{55}} & \textcolor{darkred}{\ding{55}}\\ 
    UltraEdit~\cite{zhao2024ultraedit}       & 4M           & \textcolor{darkgreen}{\ding{51}} & \textcolor{darkred}{\ding{55}}          & \textcolor{darkred}{\ding{55}} & \textcolor{darkred}{\ding{55}}\\
    AURORA~\cite{krojer2024learning}          & 280K   & \textcolor{darkgreen}{\ding{51}}      & \textcolor{darkgreen}{\ding{51}}         & \textcolor{darkred}{\ding{55}} & \textcolor{darkred}{\ding{55}}\\
    IER~\cite{tan2019expressing} & 4K & \textcolor{darkred}{\ding{55}} & \textcolor{darkgreen}{\ding{51}}     & \textcolor{darkgreen}{\ding{51}} & \textcolor{darkgreen}{\ding{51}}\\
    GIER~\cite{shi2020benchmark}            & 6K           & \textcolor{darkred}{\ding{55}} & \textcolor{darkgreen}{\ding{51}}         & \textcolor{darkgreen}{\ding{51}} & \textcolor{darkgreen}{\ding{51}}\\
    \midrule
    \textbf{RealEdit (Ours)}   & \textbf{57K} & \textbf{\textcolor{darkgreen}{\ding{51}}} & \textbf{\textcolor{darkgreen}{\ding{51}}} & \textbf{\textcolor{darkgreen}{\ding{51}}} & \textbf{\textcolor{darkgreen}{\ding{51}}} \\
    
    \bottomrule
    \end{tabular}%
    }
\end{table}

%% file: sec/04_analysis.tex
\section{\RealEdit dataset analysis}
\label{sec:dataset_analysis}
% We looked at requests submitted on reddit
% We decided to analyze what kind of requests they are
% What did we use?
% gpt 4o to taxonomize promtps
% Here is what we noticed people care about.
% We see that models are unable to perform these requests. 
% some tasks are impossible right now. Like make a vector. Resizing is imporssible for many models too. 

\begin{figure}[t]
    \centering
    \includegraphics[width=\linewidth]{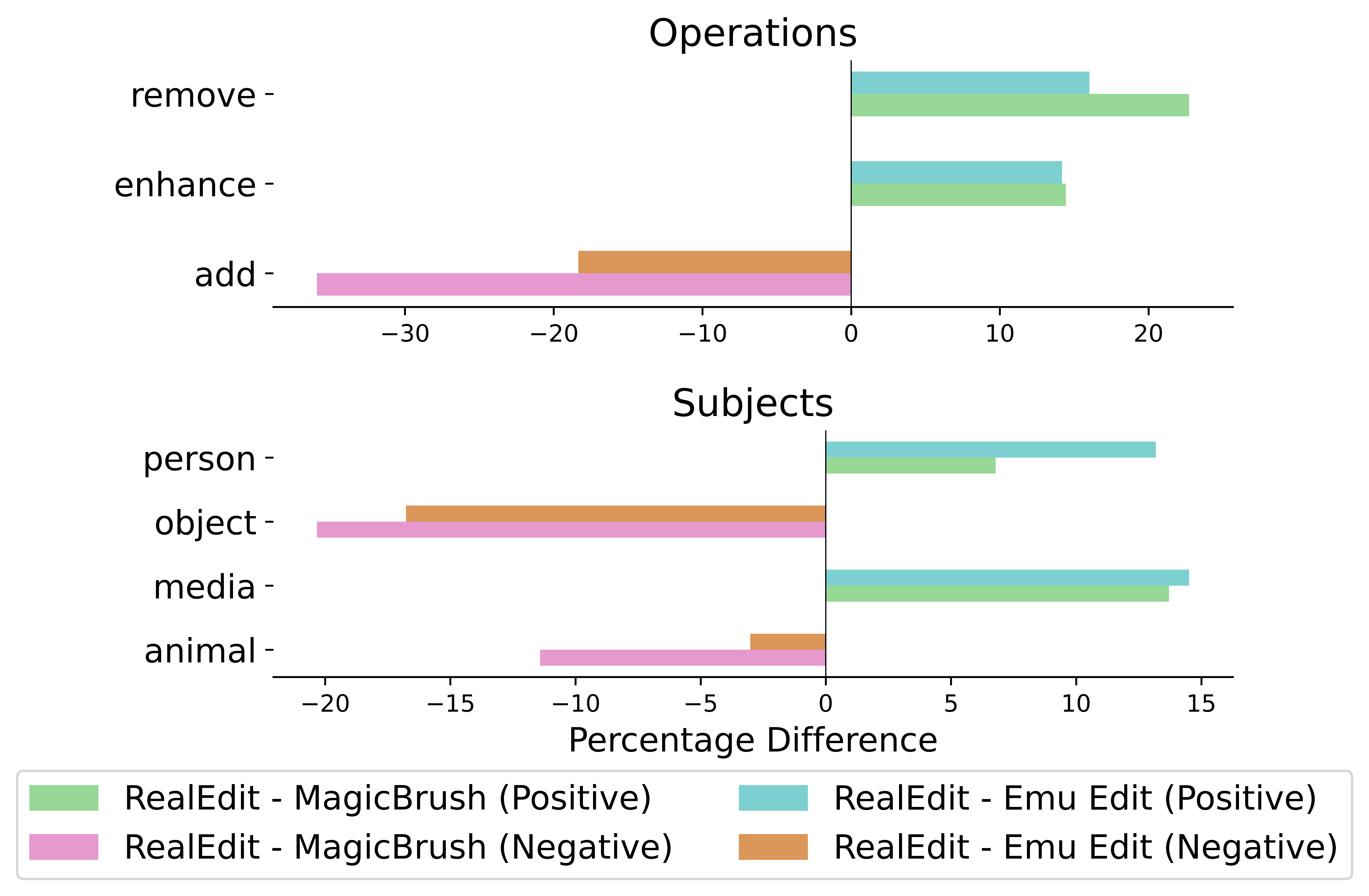}
    \caption{\textbf{Key differences in the distribution} of our test set compared to MagicBrush and Emu Edit test sets. MagicBrush and Emu Edit tend to be similar in distribution to each other, but starkly different from \RealEdit.}
    \label{fig:dist_us_both}
    \vspace{-4mm}
\end{figure}

\RealEdit provides insight into practical applications of image editing by analyzing real-world requests. We observe notable differences between \RealEdit and existing datasets including InstructPix2Pix~\cite{brooks2023instructpix2pix}, MagicBrush~\cite{zhang2024magicbrush}, Emu Edit~\cite{sheynin2024emu}, HIVE~\cite{zhang2024hive}, Ultra Edit~\cite{zhao2024ultraedit}, AURORA~\cite{krojer2024learning}, Image Editing Request~\cite{tan2019expressing} and GIER~\cite{shi2020benchmark}. 
While we focus primarily on differences with MagicBrush~\cite{zhang2024magicbrush} and Emu Edit~\cite{sheynin2024emu} in the following discussions, these observations broadly apply across datasets used to train image editing models. Figure~\ref{fig:dist_us_both} details the main differences. %We summarize the most important distinctions below.

% \begin{figure}[htbp]
%     \centering
%     \includegraphics[width=\linewidth]{figs/taxonomy_ours_fixed_font.png}
%     \caption{Taxonomy of image edit requests in our dataset. There is a wode variety of task types and edit subjects, with subtle tasks like ``remove'' and ``enhance'' being the most requested. \ranjay{I would be ok with moving this to the supplementary if we need more space.}}
%     \label{fig:taxonomy_ours}
% \end{figure}

\paragraph{Qualitative analysis and taxonomy.} %We qualitatively determine the primary subject in a sample of 500 input images from \RealEdit. 
We create a taxonomy of image editing tasks people have requested. This involves (1) categorizing our edit requests into \textit{operations}, (2) subcategorizing requests by \textit{subject} of the edit, and (3) prompting GPT-4o~\cite{openai2023gpt4} to categorize the request based on the input image and edit instruction. To determine the operations, we modify the MagicBrush~\cite{zhang2024magicbrush} set of operations for clarity. We base our possible subjects on the significant categories from the sample of 500 images. We tune our GPT-4o prompt using samples of 100 data points to ensure accuracy, then validate on a separate sample to avoid overfitting. We find that both the categories and their distribution differ greatly from prior work. Since our categorizations are fairly similar to MagicBrush~\cite{zhang2024magicbrush} and Emu Edit~\cite{sheynin2024emu} test sets, we run our taxonomy on these test sets and highlight key distributional differences in Figure~\ref{fig:dist_us_both}. Full taxonomies and comparisons are listed in Appendix A. 
% \ref{supp:taxonomy}.

\paragraph{Differences in edit operations.}
Synthetic datasets contain a greater use of ``add'' requests (36\% less than MagicBrush). In contrast, real-life photos typically contain the intended objects within the frame, with many of \RealEdit's semantically focused tasks involving the \textit{removal} of unintended elements, such as strangers in the background, shadows on faces, or cars on the street.
% \RealEdit contains a larger proportion of ``remove'' requests (diff. of 23\% MagicBrush and 16\% Emu Edit).
Additionally, there are numerous cases where input images are semantically aligned with the owner's intent, but contain errors in photography such as bad lighting, motion blur, or graininess.
Following this, \RealEdit contains more ``enhance'' requests (14\% greater than MagicBrush and Emu Edit) compared to existing datasets. These findings indicate that real users often prioritize \textit{subtler requests}, whereas synthetic datasets are dominated by larger semantic changes, such as ``add."

\paragraph{Differences in image content.}  Analysis on 500 samples reveals that around 55\% of the input images feature \textit{people} as the main subject. Consequently, the subjects of the requested edits are more likely to be people (13\% more than Emu Edit), and less likely to be man-made objects (20\% less than MagicBrush).
Animals and media (characters, movie/book posters, memes, etc.) are the next most common categories, comprising about 10\% of the test set each. Common media requests include restoring old photographs, participating in fandoms, making memes, or other forms of online entertainment.
The fixation on media is not paralleled in other datasets (15\% more than Emu Edit). 
These findings reveal a clear difference: Reddit users tend to prioritize \textit{personal significance} by including people and \textit{entertainment} by incorporating media. Synthetic datasets often fail to reflect these preferences accurately.

Given the substantial distributional differences of \RealEdit compared to existing datasets, we demonstrate in Section~\ref{sec:results} that current models struggle to perform well on real-world requests.

% \noindent\textbf{Additional considerations.} In \RealEdit, approximately 36\% of images are 1080p, indicating that humans are more concerned with editing higher resolution images.  Instructions exceed 77 tokens in approximately 3.4\% of cases. Editing models should cater to such preferences in resolution and instruction length in order to better serve human users.

% Given the substantial distributional differences of \RealEdit compared to existing datasets, we demonstrate in Section \ref{sec:results} that current models struggle to perform well on editing tasks within the \RealEdit dataset.

% Considering the significant differences in distribution of \RealEdit compared to existing models, we will prove in Section \ref{sec:results} that existing models cannot perform well on editing tasks in the \RealEdit dataset.

%% file: sec/05_method.tex
\section{An editing model trained with \RealEdit}
To demonstrate the value of \RealEdit, we develop an image editing model using training examples in \RealEdit.
Specifically, we utilize InstructPix2Pix~\cite{brooks2023instructpix2pix} as the model backbone on which we finetune using our data.
We leave exploration of different base models for future work.

\input{tables/our_benchmark_all_empty_column}

\paragraph{Aligning with pretraining data.}
Since we finetune InstructPix2Pix rather than training the model from scratch, we align our finetuning dataset with the data distribution used in InstructPix2Pix's pretraining data to avoid substantial distributional shifts that may deteriorate model's performance.
In particular, InstructPix2Pix~\cite{brooks2023instructpix2pix} applies CLIP-based~\cite{radford2021learning} filtering to ensure the quality of image pairs. In addition, as it employs Prompt-to-Prompt~\cite{hertz2022prompt} in generating its training data, the input-output image pairs are with high structural similarity.
To align our training set, we thus follow the same CLIP-based filtering and additionally use SSIM~\cite{wang2004image} to include structurally similar images, recognizing that human edits collected in \RealEdit often alter structure with techniques like drag-and-drop adjustments and symmetrical flipping.
Tasks incompatible with InstructPix2Pix’s capabilities, such as resizing images, changing file types, or highly ambiguous prompts (particularly those involving humor) are thus excluded.
In total, we trained on 39K examples.
Aligning our training data with the InstructPix2Pix distribution allows for more competitive performance on metrics, and accounts for limitations in the InstructPix2Pix's architecture and pretraining.
For training our model, we closely follow the configuration of MagicBrush~\cite{zhang2024magicbrush}. Specifically, we train our model for 51 epochs, utilizing cosine learning rate decay and incorporating a learning rate warm-up phase (details in Appendix E.
% \ref{supp:imp_details}).
\paragraph{Decoding at inference.}
We observe that Stable Diffusion~\cite{rombach2022high} struggles with accurately reconstructing human faces and fine-grained image details. As shown in Section~\ref{sec:dataset_analysis}, real-world requests are human-centric with detailed edit needs. To address this, we incorporate OpenAI's Consistency Decoder~\cite{openai_consistencydecoder} at inference time, significantly enhancing generation quality for faces, patterns, and text without altering the diffusion process.

% For training, we adhered closely to InstructPix2Pix’s~\cite{brooks2023instructpix2pix} methodology, with our primary modification being the addition of cosine learning rate decay. Full configuration details will be shared for reproducibility.

% For data selection, we applied the following criteria:
% - SSIM above 0.6
% - CLIP score above 0.7, following IP2P's approach
% - CLIP score below 0.99 to eliminate duplicates
% - Exclusion of creative prompts to reduce ambiguity
% - Removal of formatting tasks due to execution challenges

%% file: tables/our_benchmark_all_empty_column.tex
\begin{table*}[t!]
    \centering
    \caption{\textbf{Quantitative evaluation} on \ours test set. On all metrics other than pixel distance, the \ours model scores the highest.}
    \label{tab:our_benchmark_quantitative}
    \resizebox{\linewidth}{!}{%
    \begin{tabular}{lcccccc p{0.01cm} ccccc}
    \toprule
    \textbf{Model} & \textbf{VIES\_O $\uparrow$} & \textbf{VIE\_PQ $\uparrow$} & \textbf{VIE\_SC $\uparrow$} & \textbf{VQA\_llava $\uparrow$} & \textbf{VQA\_Flan-t5 $\uparrow$} & \textbf{TIFA $\uparrow$} & & \textbf{L1 $\downarrow$} & \textbf{L2 $\downarrow$} & \textbf{CLIP-I $\uparrow$} & \textbf{DINO-I $\uparrow$} & \textbf{CLIP-T $\uparrow$} \\
    \cmidrule(lr){1-7} \cmidrule(lr){9-13}
    AURORA~\cite{krojer2024learning}            & 2.20 & 3.43 & 2.40 & 0.606 & 0.711 & 0.724 & & 0.154 & 0.069 & 0.793 & 0.733 & 0.246 \\
    HIVE~\cite{zhang2024hive}                   & 1.73 & 3.40 & 1.86 & 0.596 & 0.678 & 0.685 & & 0.246 & 0.142 & 0.743 & 0.646 & 0.250 \\

    % Edit Friendly & 2.93  & \textbf{5.68} & 3.00 & & & & & 0.199 & 0.091 & 0.762 & 0.736 & \textbf{0.261} \\
    InstructPix2Pix~\cite{brooks2023instructpix2pix} & 1.64 & 3.12 & 1.76 & 0.594 & 0.650 & 0.698 & & 0.181 & 0.073 & 0.752 & 0.638 & 0.244 \\
    MagicBrush~\cite{zhang2024magicbrush}      & 1.87 & 3.88 & 1.89 & 0.620 & 0.726 & 0.741 & & \textbf{0.138} & \textbf{0.064} & 0.830 & 0.782 & 0.251 \\
    Null-text Inv.~\cite{mokady2023null}       & 1.89 & 3.27 & 2.14 & 0.637 & 0.751 & 0.731 & & 0.152 & 0.067 & 0.743 & 0.669 & \textbf{0.261} \\
    SDEdit~\cite{meng2021sdedit}              & 0.59 & 1.47 & 0.75 & 0.588 & 0.653 & 0.703 & & 0.156 & 0.068 & 0.678 & 0.613 & 0.230 \\
    % Turbo-Edit & 2.80 & 5.19 & 2.73 & & & & & 0.147 & 0.066 & 0.782 & 0.739 & \textbf{0.261}\\
    \textbf{RealEdit (Ours)}                   & \textbf{3.68} & 4.01 & \textbf{4.61} & \textbf{0.660} & \textbf{0.795} & \textbf{0.751} & & 0.143 & 0.066 & \textbf{0.840} & \textbf{0.792} & \textbf{0.261} \\

    \bottomrule
    \end{tabular}%
    }
\end{table*}

%% file: sec/06_experiments.tex
\section{Experiments} \label{sec:results}
% We evaluate our model trained on \RealEdit across various evaluation benchmarks, where  model demonstrates strong performance on both real user data and synthetic datasets.
\noindent\textbf{Setup.} We benchmark our model against six open-source baselines: InstructPix2Pix~\cite{brooks2023instructpix2pix}, MagicBrush~\cite{zhang2024magicbrush}, AURORA~\cite{krojer2024learning}, SDEdit~\cite{meng2021sdedit}, HIVE~\cite{zhang2024hive}, and Null-text Inversion~\cite{mokady2023null}.
We leverage the input and output captions generated in Section~\ref{sec:realedit} for models that require them.

% LLaVA-Next and GPT-4o as mentioned  

To evaluate the models, we adopt a comprehensive suite of metrics.
First, we utilize VQA-based automated metrics to measure task completion, as these metrics have been shown to closely reflect human judgments. In particular, we use VIEScore~\cite{ku2023viescore} with a GPT-4o backbone as our default metric, as it evaluates semantic consistency (VIE\_SC), perceptual quality (VIE\_PQ), and overall alignment with human-requested edits (VIE\_O) each on a scale of 0 to 10. 
Similarly, we use VQAscore~\cite{lin2024evaluating} (with different base models: 
 LLaVa and FLAN-T5) and TIFA~\cite{hu2023tifa} to evaluate the fine-grained faithfulness of the output image to the edit instruction.
We also include standard metrics such as L1- and L2 pixel distance, DINO~\cite{zhang2022dino}, CLIP-I and CLIP-T, following prior work~\cite{zhang2024magicbrush, sheynin2024emu}.
Most importantly, we leverage real users to make pairwise comparisons between edits and compute Elo ranking of the models~\cite{jiang2024genai}. We further qualitatively study the response Reddit users have on edits produced by our model on recent posts.
% \noindent\textbf{Training configuration.}

\subsection{Automated evaluations on \ours test set}
In Table~\ref{tab:our_benchmark_quantitative}, we show that existing models struggle to capture the semantic nuances of human requests, while our model achieves notable improvements, particularly in VIE\_SC scores. Our model also significantly outperforms other baselines on finer-grained metrics like VQAScore and TIFA. Although our model achieves state-of-the-art (SOTA) results on standard metrics, these metrics are limited as they fail to fully capture task completion. Notably, using the input image as the output yields the highest scores on four out of five metrics, with the fifth, CLIP-T, exhibiting saturation effects. This underscores the importance of more nuanced automated metrics, such as VQA-based approaches, to better evaluate task completion.

% These additional metrics indicate that our model’s performance is robust and capable of generalizing beyond its proprietary data, demonstrating it is not simply overfitting.

\subsection{Human evaluation on \ours test set}

Methods like VIEScore\cite{ku2023viescore} align more closely with human judgment, but rely on vision-language models, which often miss subtle differences and produce inconsistent results. 
%Thus, human judgment remains the most dependable evaluation method.

\input{tables/ELO.tex}

% \noindent\textbf{Elo score with human preference.}
To counteract this, we conducted a qualitative evaluation using Elo scores, following the methodology from GenAI Arena~\cite{jiang2024genai} and LMSYS~\cite{zheng2023judging}. This evaluation, conducted via Amazon Mechanical Turk, involved pairwise comparisons against the baselines on 200 diverse images from our test set. Results in Table~\ref{tab:elo_rankings} demonstrate that our model outperforms baselines on human judgement.

% \ranjay{I elevated the reddit experiments to be their own subsection.}
\subsection{Deploying our model on Reddit}
One limitation of standard Elo evaluations is that they are conducted by individuals with no personal connection to the image. To ecologically validate the utility of our model with photo owners, we deploy our model back on Reddit. We provide editing services for new user requests, posting edited images in the comments per subreddit guidelines.

On multiple occasions, we received positive feedback.
For example, the model successfully removed red-eye from a photo. The original poster (OP) responded with: ``Thank you so much! Solved." and closed the request. On another occasion, we edited a picture of a car, and the OP remarked, ``It looks pretty good, man." On an edit of removing a person from the background, OP commented ``Wow this looks great! I love the way you smoothed out the lighting on me as well'' indicating that our model not only is successful semantically but produces aesthetic images.

\begin{figure}[!h]
    \centering
    \includegraphics[width=\linewidth]{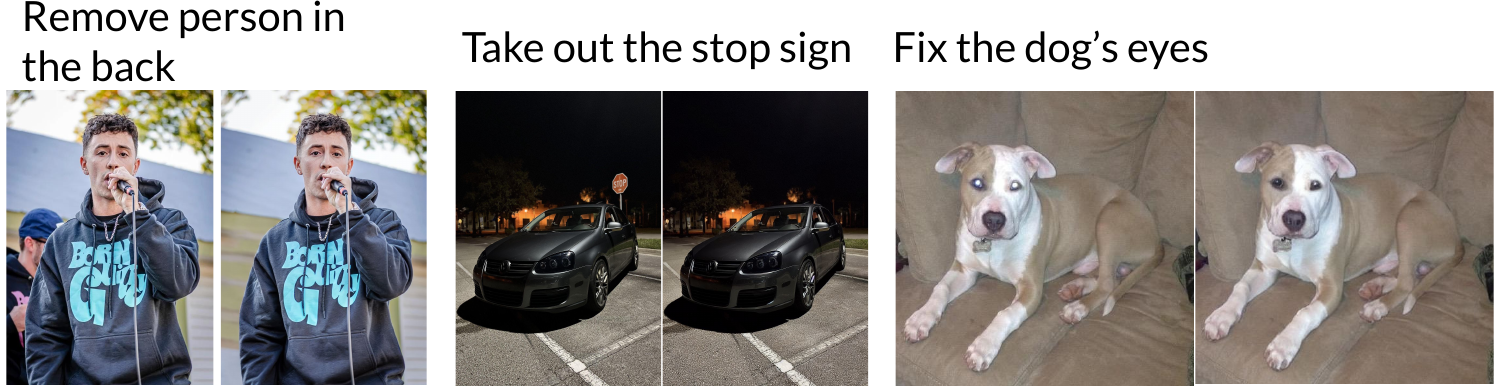}
    \caption{\textbf{Real requests completed on Reddit.} We deployed our model on r/PhotoshopRequest to complete in-the-wild requests. We received positive feedback from users on the examples above.}
    \label{fig:reddit_figure}
    \vspace{-2mm}
\end{figure}

% The best measure of model performance is whether the request issuer is satisfied with the result. Different people often have different tastes for subjective editing tasks, and the image owner has the greatest stake and most relevant perspective on the edit’s success. To assess this, we turn to the original inspiration for this work: the Reddit community. We collect recent requests from r/estoration (and possibly r/PhotoshopRequests) and qualitatively analyze responses from other commenters as well as the original poster (OP). We find that [INSERT RESULTS HERE].

\subsection{Evaluations on existing test sets}

We also conduct evaluations on external test sets including the test sets in GenAI Arena~\cite{jiang2024genai}, Emu Edit~\cite{sheynin2024emu}, and MagicBrush~\cite{zhang2024magicbrush}.
On GenAI Arena, we report Elo ranking in Table~\ref{tab:elo_rankings} computed with real human preferences.
Our model ranks second among the evaluated models. While these results were insightful, we found the examples in GenAI Arena to be less representative of real-world tasks. 
% Additionally, it includes MagicBrush data, which is an inherent bias. 
Additionally, since GenAI Arena’s data comes from ImagenHub, which includes MagicBrush data, the evaluation is inherently biased toward MagicBrush.
%For instance, the first test case shows a picture of a zebra and has the instruction of: ``Give the zebra a single front leg''.
We include full automated evaluation results on Emu Edit and MagicBrush in Appendix F,
% \ref{supp:ood_eval},
where our model performs competitively with the individual strongest models on respective test sets across varying metrics.
% We use the CLIP-based metrics proposed in their papers, with detailed results available in Appendix.
% The Emu Edit metrics are particularly limited, as they primarily measure similarity to the input image rather than ground truth, which does not accurately reflect task completion.

% We evaluate our model on the Emu Edit~\cite{sheynin2024emu} and MagicBrush~\cite{zhang2024magicbrush} test sets.
% We use the CLIP-based metrics proposed in their papers, with detailed results available in Appendix. The Emu Edit metrics are particularly limited, as they primarily measure similarity to the input image rather than ground truth, which does not accurately reflect task completion.
%\input{tables/mb_emu_metrics}

\subsection{Improving edited image detection}

\input{tables/truemedia_new}
\begin{figure}[!t]
\label{paris}
\centering\includegraphics[width=0.38\textwidth]{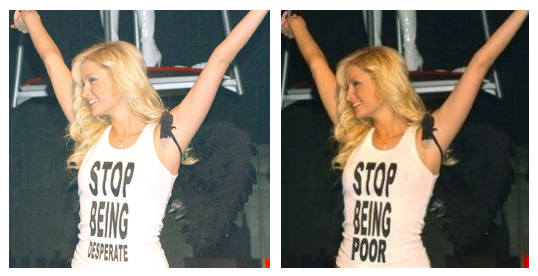}
\caption{The baseline misclassifies both images as real, whereas our model correctly spots the fake (right) that spawned the 2005 Paris Hilton ``Stop Being Poor'' meme.}
\end{figure}

%Here, we discuss edited image detection as another use case of \RealEdit data.
We partnered with \truemedia, a platform where users can upload media to assess authenticity. Their primary fake image detection model is a fine-tuned version of Universal Fake Detect (UFD) ~\cite{ojha2024universalfakeimagedetectors}, which effectively detects model-generated deepfakes. 
%While effective at detecting model-generated deepfake images, UFD fails to reliably detect \textit{human-edited} images. 
We leverage the human-edited images in \RealEdit to enhance the model’s ability to detect such edits, which has significant real-world impact. 

UFD is trained on a recipe of 62K images from academic datasets~\cite{wang2023diffusiondblargescalepromptgallery, karras2019style, karras2020analyzingimprovingimagequality, karras2018progressivegrowinggansimproved, lin2015microsoftcococommonobjects, sdfd} and some proprietary data, none of which includes \textit{human edits}. We trained a model from scratch using the UFD training pipeline with added \RealEdit training data. We evaluated on the \RealEdit test set and on a random subset of 100 reals plus 100 in-the-wild edited images from \truemedia.
% less than 20\% of which involved AI-manipulation.
We show in Table~\ref{tab:ufd-combined} that fine-tuning on \RealEdit improves F1 by 45.5 and 14 points on \RealEdit and \truemedia's test sets respectively.
% This is likely because \truemedia’s model had limited exposure to manually edited images.
\RealEdit also serves as a challenging human-edited image detection benchmark for models that are more specialized for this task compared to deepfake detection~\cite{triaridis2024mmfusioncombiningimageforensic, zhang2023editguardversatileimagewatermarking, 6625374, Zhang_2024}.

% Fine-tuning UFD on \RealEdit improves F1 by 14\%, but there remains significant room for improvement. This highlights the difficulty of detecting in-the-wild edits, which often lack the traceable artifacts found in fully AI-generated images.

% Grad-CAM result would go here
% To understand how our model learned, we computed the saliency map of the above example using Grad-CAM \cite{Selvaraju_2019} and find that 

% \begin{figure}[H]
% \label{saliency}
% \centering\includegraphics[width=0.35\textwidth]{figs/edit_detection.PNG}
%     \caption{Saliency map of the Paris Hilton image computed with Grad-CAM}
% \end{figure}

\begin{figure*}[!t]
    \centering
    \includegraphics[width=\textwidth]{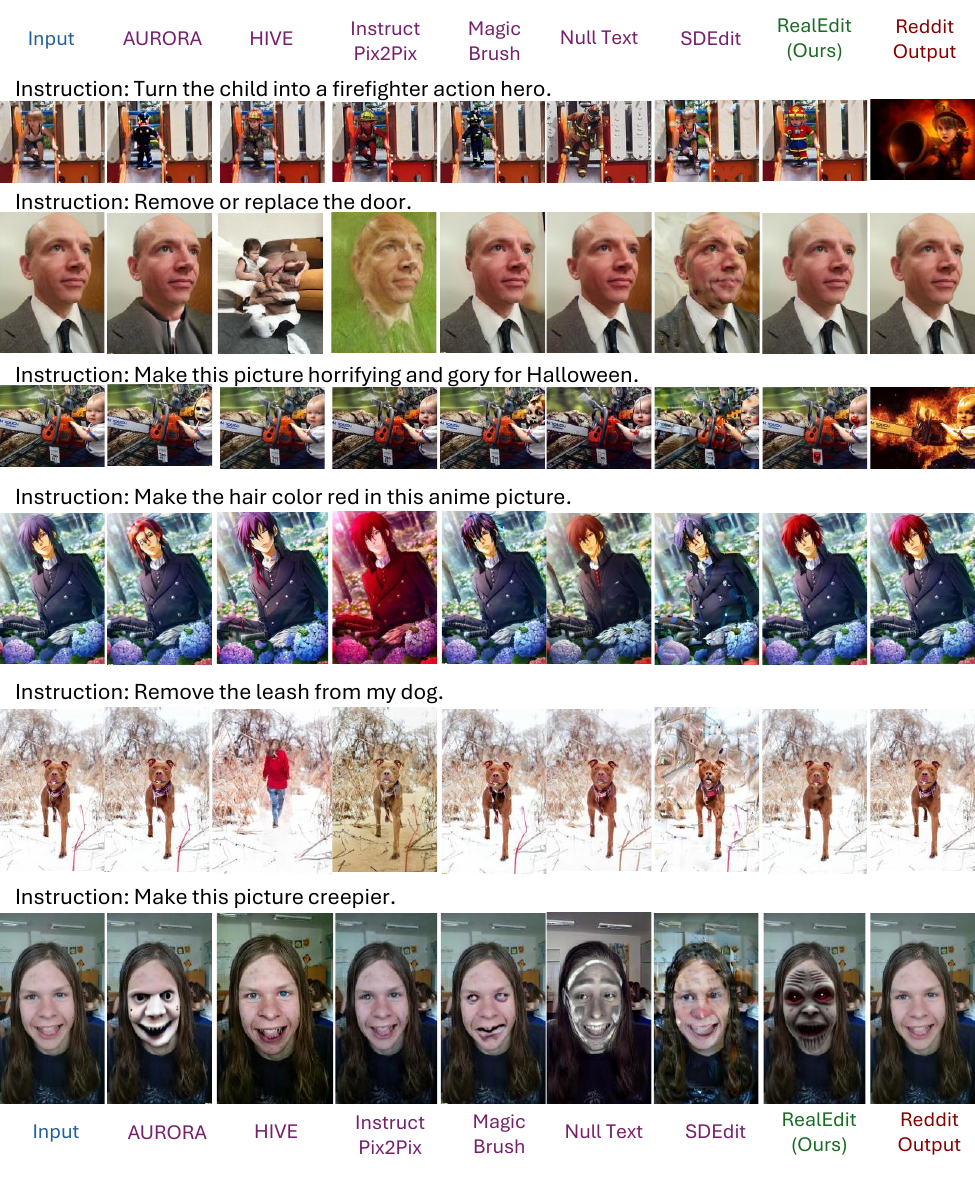}
    \caption{\textbf{Examples of the \RealEdit model} on \RealEdit test set images compared to other editing models. Our edits are often more semantically correct as well as more visually appealing.}
    \label{fig:examples}
\end{figure*}

%% file: tables/ELO.tex
% \setlength{\tabcolsep}{1pt}
\begin{table}[t]
    \centering
    \caption{\textbf{Elo rankings} on \textsc{RealEdit} and GenAI Arena~\cite{jiang2024genai} test sets. On \RealEdit test set, our model scores the highest. We perform competitively on GenAI Arena test set of synthetic data.}
    \label{tab:elo_rankings}
    % \resizebox{\linewidth}{!}
    \begin{tabular}{lcc | cc}
    \toprule
    \multirow{2}{*}{Model}& \multicolumn{2}{c|}{\textsc{RealEdit}} & \multicolumn{2}{c}{\textsc{GenAI Arena}} \\
    \cmidrule{2-5}
     & Elo & 95\% CI & Elo & 95\% CI \\
    \midrule
    AURORA~\cite{krojer2024learning} & 1019 & +14/-11 & - & - \\
    HIVE~\cite{zhang2024hive} & 997 & +16/-10 & - & - \\
    InstructPix2Pix~\cite{brooks2023instructpix2pix} & 984 & +13/-16 & 1011 & -50/+47 \\
    MagicBrush~\cite{zhang2024magicbrush} & 982 & +11/-13 & \textbf{1107} & -39/+47 \\
    Null-text Inv.~\cite{mokady2023null} & 949 & +10/-11 & - & - \\
    SDEdit~\cite{meng2021sdedit} & 885 & +13/-13 & 991 & -48/+35 \\
    \textbf{RealEdit (Ours)} & \textbf{1184} & +17/-12 &  1043 & -12/+17 \\
    % COSXLEdit~\cite{CosXL} & - & - & 1063 & -49/+42 \\
    % InfEdit~\cite{xu2023infedit} & - & - & 1023 & -44/+39 \\
    % Prompt2prompt~\cite{hertz2022prompt} & - & - & 1010 & -46/+46 \\
    % PNP~\cite{Tumanyan_2023_CVPR} & - & - & 991 & -43/+62  \\
    % Cycle Diffusion~\cite{cyclediffusion} & - & - & 932 & -41/+49 \\
    % Pix2PixZero~\cite{parmar2023zero} & - & - & 833 & -46/+41 \\
    \bottomrule
    \end{tabular}%
    % }
    \vspace{-0.1cm}
\end{table}

%% file: tables/truemedia_new.tex
\begin{table}[t]
% \caption{Model performance on detecting edited images on \ours test set and in-the-wild user uploaded images to \truemedia. Our model increases F1 on the in-the-wild images by 14\%.}
\caption{Model performance on detecting edited images in the \ours test set and in-the-wild images on \truemedia.}
\label{tab:ufd-combined}
\centering
\small
\resizebox{\linewidth}{!}{%
\begin{tabular}{l ccc|ccc}
\toprule

& \multicolumn{3}{c|}{\textbf{\ours dataset}} & \multicolumn{3}{c}{\textbf{In-the-Wild dataset}} \\
\midrule
\textbf{Model} & \textbf{F1}$\uparrow$ & \textbf{Recall}$\uparrow$ & \textbf{Precision}$\uparrow$ & \textbf{F1}$\uparrow$ & \textbf{Recall}$\uparrow$ & \textbf{Precision}$\uparrow$  \\
\midrule
Baseline & 23.5 & 14 & \textbf{80} & 49 & 35 & \textbf{80} \\
\textbf{Ours} & \bf{69} & \bf{64} & 74 & \bf{63} & \bf{57} & 71 \\
\midrule
Change & + 45.5 & +50 & -6 & +14 & +22 & -9 \\
\bottomrule
\end{tabular}%
    }
\end{table}

%% file: sec/07_discussion.tex
\section{Discussion}

\paragraph{Privacy and ethics.}

% To protect user privacy, individuals
% can opt out of having their images in the dataset by removing the photos from Reddit. Since our dataset contains image URLs rather than image files, images deleted from the
% web are automatically removed. Additionally, we provide a
% form where individuals can request their data to be removed
% from the dataset. Although this evolving dataset may introduce challenges for quantitative validation, ensuring user
% privacy remains our top priority.
% Our work has positive social impacts, such as reducing
% the need for professional editing software and skills, and
% enabling higher-quality restorations of family photographs.
% However, we recognize the risks of malicious exploitation
% and strongly oppose any harmful, offensive, or derogatory
% use of our model or data. We plan to further pursue the
% development of fake image detection tools.

To protect user privacy, individuals can opt out of having their images in the dataset by removing the photos from Reddit. Since our dataset contains image URLs rather than image files, images deleted from the web are automatically removed. Additionally, we provide a form where individuals can request their data to be removed from the dataset. 
% Although this evolving dataset may introduce challenges for quantitative validation, ensuring user privacy remains our top priority. 

Our work has positive social impacts, such as reducing
the need for professional editing software and skills, and
enabling higher-quality restorations of family photographs.
However, we recognize the risks of malicious exploitation
and strongly oppose any harmful, offensive, or derogatory
use of our model or data.

\paragraph{Acknowledgments}
We thank Galen Weld, Matthew Wallingford, Vivek Ramanujan, and Benlin Liu for their time, guidance, and mentorship. We additionally thank the UW RAIVN lab for support throughout the project. 

%% file: sec/X_suppl.tex
% \documentclass[10pt,twocolumn,letterpaper]{article}
% \usepackage{}

% \begin{document}

\appendix

\clearpage
\onecolumn
\noindent\rule{\textwidth}{1pt}
\begin{center}
\vspace{3pt}
{\LARGE\textbf{\textsc{RealEdit}: Reddit Edits As a Large-scale\\ Empirical Dataset for Image Transformations}}\\
\vspace{1em}
{\Large\textbf{Supplementary Material}}
\vspace{-3pt}
\end{center}
\noindent\rule{\textwidth}{1pt}

\vspace{1em}

\renewcommand{\contentsname}{Table of Contents} % title for TOC
\tableofcontents

\clearpage
\twocolumn

% \onecolumn
% \noindent\rule{\textwidth}{1pt}
% \begin{center}
% \vspace{3pt}
% {\LARGE\textbf{\textsc{RealEdit}: Reddit Edits As a Large-scale\\ Empirical Dataset for Image Transformations}}\\
% \vspace{1em}
% {\Large\textbf{Supplementary Material}}
% \vspace{-3pt}
% \end{center}
% \noindent\rule{\textwidth}{1pt}
% \centering \LARGE\textbf{\textsc{RealEdit}: Reddit Edits As a Large-scale\\ Empirical Dataset for Image Transformations}\par\vspace{0.5em}\Large{Supplementary Material}\par\vspace{1em}

% \input{tables/test}
% \section*{Table of contents}
% \tableofcontents

\clearpage

\twocolumn
\section{Data taxonomy} \label{supp:taxonomy}
\subsection{Full taxonomy}
\begin{figure}[t]
    \centering
    \includegraphics[width=\linewidth]{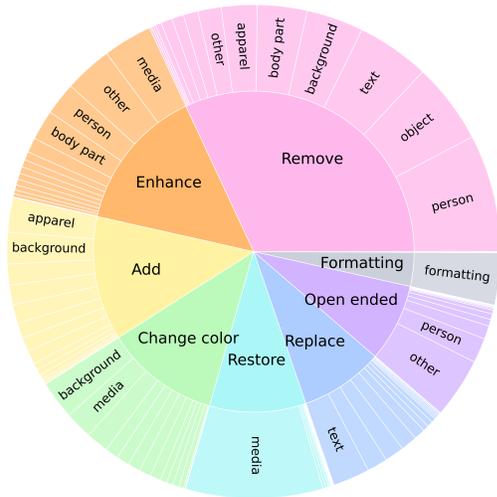}
    \caption{\textbf{Taxonomy of \ours image edit requests.} There is a wide variety of task types and edit subjects, with subtle tasks like ``remove'' and ``enhance'' being the most requested.}
    \label{fig:taxonomy_ours}
\end{figure}

\begin{figure}[htbp]
    \centering
    \includegraphics[width=\linewidth]{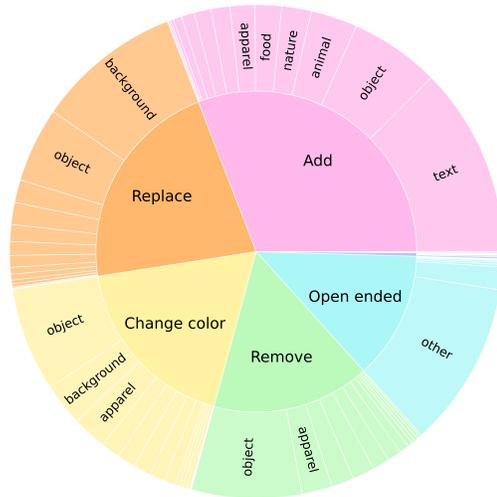}
    \caption{\textbf{Taxonomy of Emu Edit image edit requests.} There is a smaller range of task types than \ours, but the distribution is fairly even.}
    \label{fig:taxonomy_emu}
\end{figure}

\begin{figure}[htbp]
    \centering
    \includegraphics[width=\linewidth]{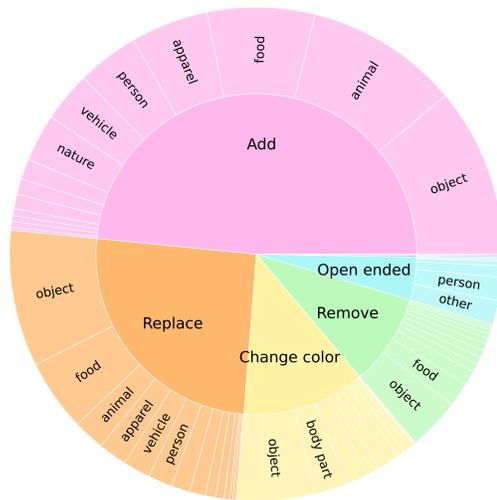}
    \caption{\textbf{Taxonomy of MagicBrush image edit requests.} There is a limited selection and extremely uneven distribution of task types, with ``add'' accounting for almost half of all requests.}
    \label{fig:taxonomy_mb}
\end{figure}

\begin{figure}[!t]
    \centering
    \includegraphics[width=\linewidth]{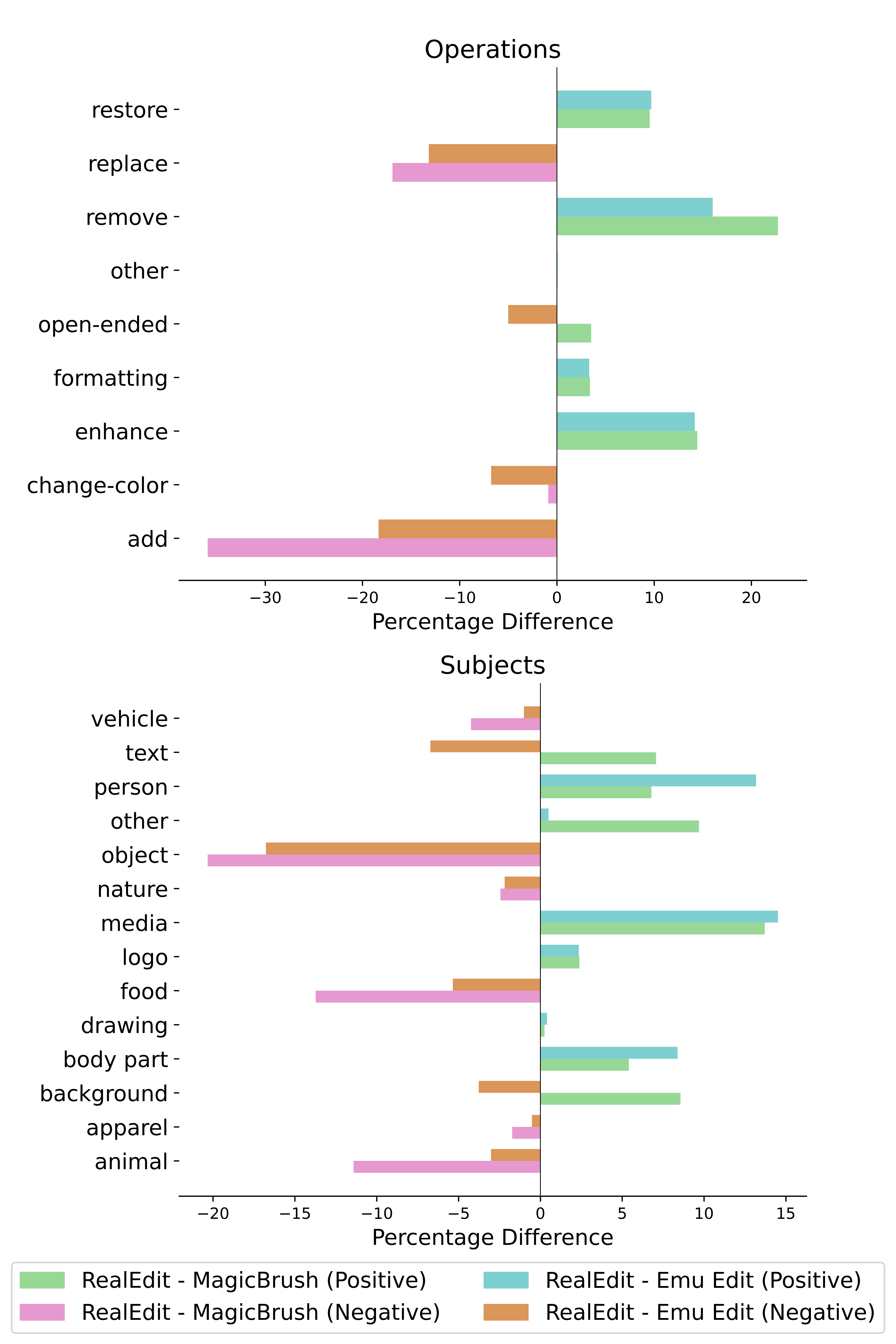}
    \caption{\textbf{Differences in the distribution} of our test set compared to MagicBrush and Emu Edit test sets. MagicBrush and Emu Edit tend to be similar in distribution to each other, but starkly different from \RealEdit.}
    \label{fig:dist_us_both_full}
    \vspace{-4mm}
\end{figure}

\begin{figure}[!t]
    \centering
    \includegraphics[width=\linewidth]{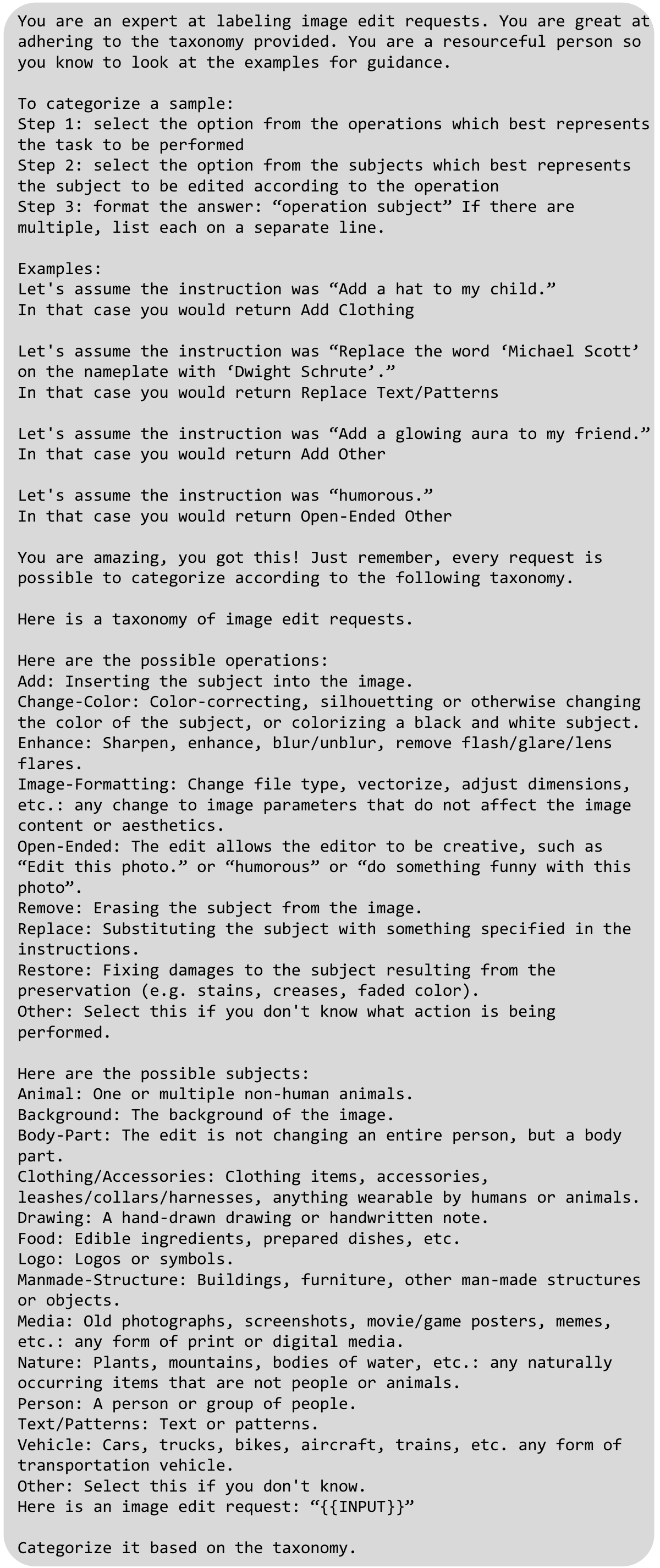}
    \caption{\textbf{Prompt used for taxonomizing edit requests.} We passed this along with input images to GPT-4o.}
    \label{fig:taxonomy_prompt}
\end{figure}

We include the taxonomies of \ours (Figure \ref{fig:taxonomy_ours}), Emu Edit (Figure \ref{fig:taxonomy_emu}), and MagicBrush (Figure \ref{fig:taxonomy_mb}) test sets, as well as the unabridged comparison between all three (Figure \ref{fig:dist_us_both_full}). The prompt used to taxonomize these requests is included in Figure \ref{fig:taxonomy_prompt}. We notice \ours has a more diverse set of tasks as well as a more even distribution with greater focus in tasks like ``remove'' and ``enhance''. Emu Edit~\cite{sheynin2024emu} has a fairly even task distribution, though a smaller set of common tasks. MagicBrush~\cite{zhang2024magicbrush} has a very skewed distribution, with a high focus on ``add'' tasks which are not likely to be requested by human users, as humans generally include all desired elements when taking a photograph. 

\begin{table}[htbp]
\centering
\begin{minipage}[t]{0.45\linewidth}
    \vspace{0pt}  % Force top alignment
    \centering
     \footnotesize
    \begin{tabular}{c|c}
    \textbf{Operation} & \textbf{\%} \\
         \hline
        Remove        & 31.9 \\
        Enhance       & 14.5 \\
        Add           & 12.5 \\
        Change Color  & 11.5 \\
        Restore       & 9.7 \\
        Replace       & 8.4 \\
        Open Ended    & 7.9 \\
        Formatting    & 3.4 \\
        Other         & 0.1 \\
    \end{tabular}
    \caption{Distribution of edit operations in the test set.}
    \label{tab:operations}
\end{minipage}
\hfill
\begin{minipage}[t]{0.45\linewidth}
    \vspace{0pt}  % Force top alignment
    \centering
    \footnotesize
    \begin{tabular}{c|c}
    \textbf{Subject} & \textbf{\%} \\
         \hline
        Person       & 15.2 \\
        Media        & 14.9 \\
        Background   & 11.0 \\
        Body Part    & 9.3 \\
        Text         & 8.9 \\
        Object       & 8.3 \\
        Apparel      & 6.5 \\
        Formatting   & 3.4 \\
        Animal       & 3.3 \\
        Logo         & 2.5 \\
        Vehicle      & 2.4 \\
        Nature       & 2.2 \\
        Other        & 12.1 \\
    \end{tabular}
    \caption{Distribution of edit subjects in the test set.}
    \label{tab:subjects}
\end{minipage}
\end{table}
\subsection{Performance across edit operations}
We show the VIEScore comparisons of \ours, AURORA~\cite{krojer2024learning}, InstructPix2Pix~\cite{brooks2023instructpix2pix} and MagicBrush~\cite{zhang2024magicbrush} in Table \ref{tab:vie_by_taxonomy}. We notice that in all of the editing tasks, the \ours model has the highest overall VIEScore. However, in ``add'' tasks, which comprise a much smaller percentage of our dataset compared to InstructPix2Pix and MagicBrush, we have a lower perceived quality, indicating that having more ``add'' data might improve the aesthetics. The task with the highest score for \ours is ``remove'', with a VIE\_O score of 4.35. The ``remove'' task comprises the largest portion of our dataset, which may explain this result. The hardest task is ``formatting'', the only operation for which we do not have the highest semantic completion score. This is due to the fact that this task is impossible for current models to fulfill properly, as changing file formats, resizing, etc. are not supported by current model architecture. 
\input{tables/vie_by_taxonomy}

% \clearpage

\section{Data processing} \label{supp:data_preprocess}
\paragraph{Test set image captioning}

We caption all input and ground truth images in the test set to enable evaluations with models that require captions. The process involves two main stages. First, for input image captioning, we pass the processed instruction along with the input image to LLaVA-Next\cite{liu2024llavanext}. This generates a caption for the input image that integrates the instruction, emphasizing key aspects of the image relevant to the editing task.

\begin{figure*}[!t]
    \centering
    \includegraphics[width=\linewidth]{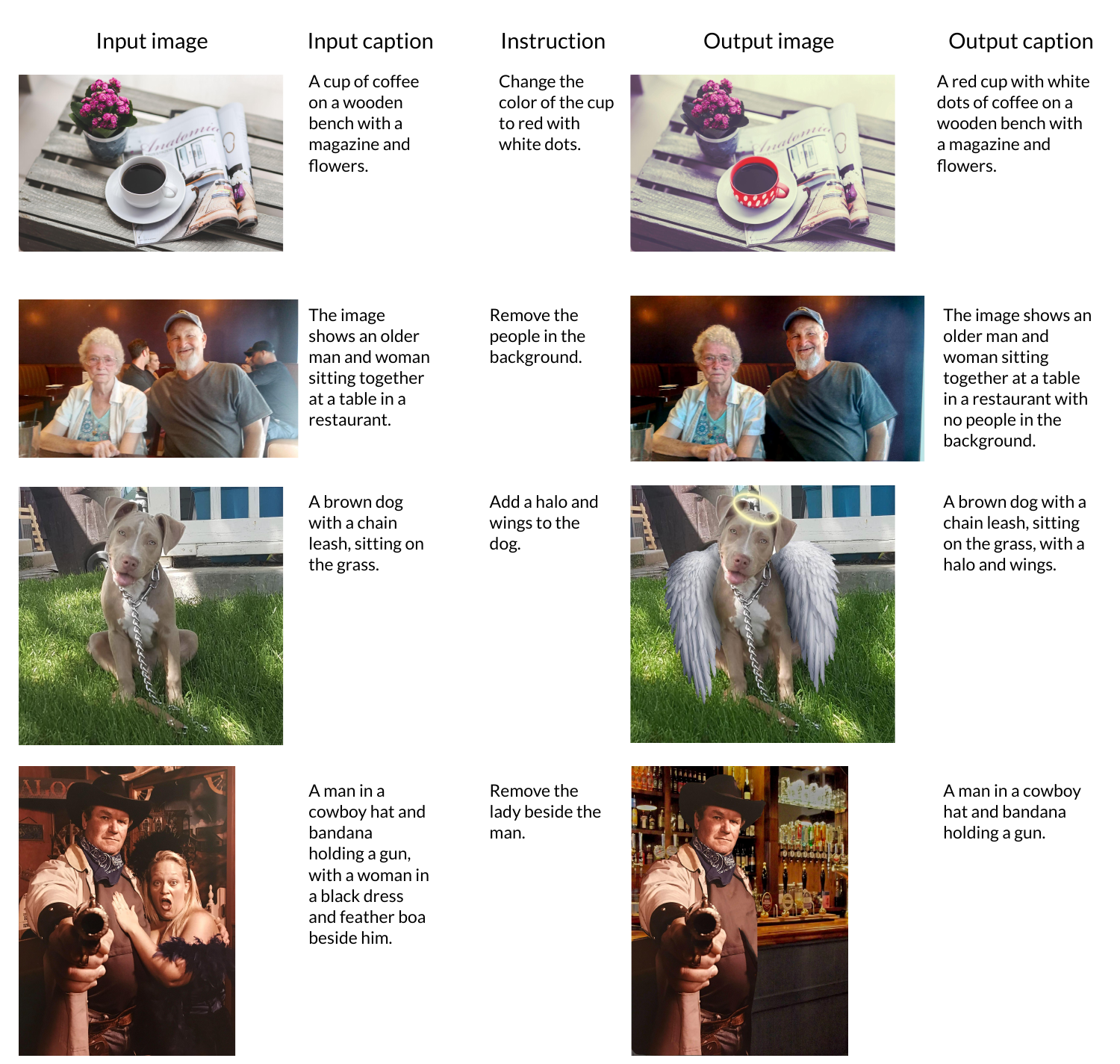}
    \caption{Examples of test set data with captions for input image and ground truth image.}
    \label{fig:test_set_figure}
\end{figure*}

For output image captioning, we pass the input caption and edit instruction to GPT-4o, which combines these elements to generate a caption for the ground truth (edited) image, reflecting both the original content and the changes made according to the instruction. Refer to Figure \ref{fig:test_set_figure} for examples of captions.

\section{Additional baselines}
After the submission of this paper, we were made aware of additional editing models with strong performance. In Table \ref{tab:additional_eval}, we compare \ours to Edit Friendly~\cite{huberman2024edit} and TurboEdit~\cite{deutch2024turboedit}.

\begin{table}[H]
    \centering
    \scriptsize
    \setlength{\tabcolsep}{3pt} % Reduce column spacing
    \renewcommand{\arraystretch}{0.85} % Reduce row spacing
    \captionsetup{skip=2pt}
    \caption{\textbf{\ours outperforms Edit Friendly and TurboEdit} on real-world edit requests.}
    \label{tab:our_benchmark_quantitative}
    \begin{tabular}{lcccccccc}
    \toprule
    \textbf{Model} & \textbf{V\textsubscript{SC}} $\uparrow$ & \textbf{V\textsubscript{PQ}} $\uparrow$ & \textbf{V\textsubscript{O}} $\uparrow$ & \textbf{L1} $\downarrow$ & \textbf{L2} $\downarrow$ & \textbf{CLIP\textsubscript{I}} $\uparrow$ & \textbf{DINO\textsubscript{I}} $\uparrow$ & \textbf{CLIP\textsubscript{T}} $\uparrow$ \\

    \midrule
    Turbo-Edit & 2.73 & 5.19 & 2.80 & 0.147 & 0.066 & 0.782 & 0.739 & \textbf{0.261}\\
    Edit Friendly & 3.00 & \textbf{5.68} & 2.93 & 0.199 & 0.091 & 0.762 & 0.736 & \textbf{0.261} \\
    \textbf{RealEdit} & \textbf{3.68} & 4.01 & \textbf{4.61} & \textbf{0.143} & \textbf{0.066} & \textbf{0.840} & \textbf{0.792} & \textbf{0.261} \\ 
    \bottomrule
    \end{tabular}
    \vspace{-0.1cm}
    \label{tab:additional_eval}
\end{table}
% Here are teh prompts.

% Here are some examples:

\section{Discussion}
\subsection{Limitations and future work}
\ours is collected from Reddit posts from 2012-2021. As such, we have less data and a danger of it getting outdated. We plan to regularly update our dataset to ensure that the edits reflect as current culture as much as possible. This will also help in edited image detection, by facilitating the detection of edits where newer AI tools were used, as the line between human editing and model editing is increasingly blurred. \\
We also filter our dataset in order to more closely match the training distribution, removing some natural diversity of human edit requests. In future work, we hope to explore different architectures capable of handling real world edit requests and editing styles. \\
The pretraining of the \ours model uses CLIP embeddings, which while very useful for semantic changes to an image, a large portion of the \ours dataset involves edits that do not involve semantic changes. Additionally, in edited image detection, some of the edits may not change the embeddings much. We urge future work to explore alternatives to such embeddings that may capture purely aesthetic changes.
\subsection{Social impacts}
The social impact of our dataset stems from both the effect on model training as well as the ability of our test set to be used to accurately and justly benchmark other models. The training data will inform how well the \ours model performs certain types of edits. The test set on the other hand determines the factors we incentivize in other models. \\
Accessible image editing models that are capable of handling real world tasks are extremely useful in democratizing the documentation of people's lives. For example, some requests in \ours involve restoring old photographs, many of which were paid. The \ours model can help more users to document meaningful family histories, even if they cannot afford to pay for edits. We have demonstrated the efficacy of our model on making real world edits by uploading our model's generations to Reddit. Additionally, our exploration of the contribution of \ours in deepfake image detection has shown that \ours increases the ability of \truemedia's ability to detect fake images, which is extremely useful in a world where images are routinely edited to cause scandals or spread misinformation. \\
There is a known issue in image generation models of generating images or making edits based on demographic biases such as smoothing wrinkles, lightening skin, and male bias in certain professions, which may offend users. Additionally, our dataset mirrors the demographic profile of Reddit users, who are predominantly Western, younger, male, and left-leaning, potentially influencing the types of images and editing requests included. We hope to study the effect of this extensively in \ours in future work.\\
There is also an issue of inappropriate edits, which we have mitigated to our knowledge in \ours through filtering of NSFW content using opennsfw~\cite{bhky_opennsfw2}, along with manual filtering in our test set. \\

\subsection{Ethics}
Some other editing datasets \cite{zhang2024magicbrush} do not use human faces in order to evade biases as well as privacy concerns. However, in \ours, we determine that since over half of edit requests contain images focused on people, we must train on human data in order to be successful in completing real world editing tasks. To mitigate privacy concerns, we use the URL in place of the actual input image so that if the original poster (OP) deletes their post, it will be removed from our dataset. We also include a form for users to request their data to be removed. This follows the standards set by RedCaps ~\cite{desai2021redcaps}. In the case of mitigating biases, we hope in future work to study the effects of using Reddit data on task completion for a wide array of demographic groups, as well as techniques or supplementary data sources to boost performance on underrepresented groups. This is a known problem in the field, and we are compelled by user preferences to include human data. Given this, although we appreciate the importance of mitigating demographic biases, this is outside the scope of a single paper.

% \clearpage

\section{Modeling ablations}

\subsection{Implementation details} \label{supp:imp_details}

We fine-tune the checkpoint of InstructPix2Pix~\cite{brooks2023instructpix2pix} using the \ours training set for 51 epochs on a single 80GB NVIDIA A100 GPU. The total batch size is 128, and the learning rate starts at $2 \times 10^{-4}$
. We resize images to 256 × 256, disable symmetrical flipping to maintain structural integrity, and apply a cosine learning rate decay to $10^{-6}$ over 15,000 steps with 400 warmup steps. The training process takes 24 hours.

\subsection{Consistency decoder}
\input{tables/consistency_decoder_metrics}

We integrate OpenAI’s Consistency Decoder~\cite{openai_consistencydecoder}, which is designed to enhance the quality of specific features during inference. This has a minimal impact on overall model performance metrics but proves highly effective for improving the handling of faces, textures, and intricate patterns. 

\begin{figure*}[t]
    \centering
    \includegraphics[width=\textwidth]{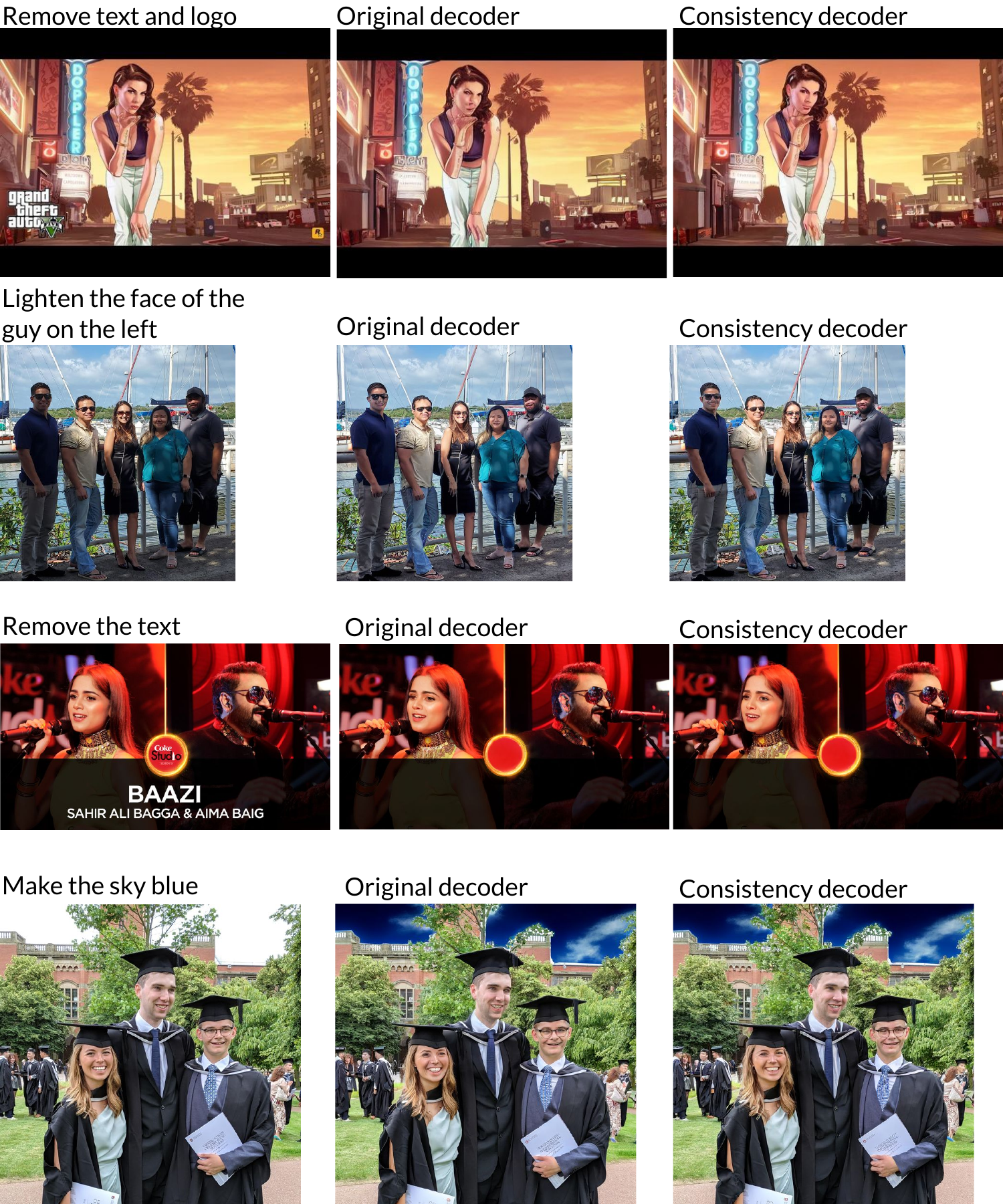}
    \caption{Consistency decoder allows for more aesthetic generation of faces.}
    \label{fig:decoder_figure}
\end{figure*}

As the decoder operates independently of the underlying model, we evaluate its effectiveness with InstructPix2Pix\cite{brooks2023instructpix2pix} and MagicBrush\cite{zhang2024magicbrush} on a sample of 500 tasks. The results indicate that while the decoder minimally affects standard metrics, such as VIEScore\cite{ku2023viescore} and CLIP-T (Table \ref{tab:consistency_decoder_combined}), it often enhances the aesthetic quality in areas requiring fine detail, such as facial reconstruction and complex textures (Figure \ref{fig:decoder_figure}).

These findings demonstrate the decoder’s potential as a lightweight, inference-only addition to improve the output quality of existing image-editing models without altering their core architectures or diffusion processes.

% \clearpage

% \clearpage

\subsection{Data filtering}

\begin{figure}[b]
    \centering
    \includegraphics[width=\columnwidth]{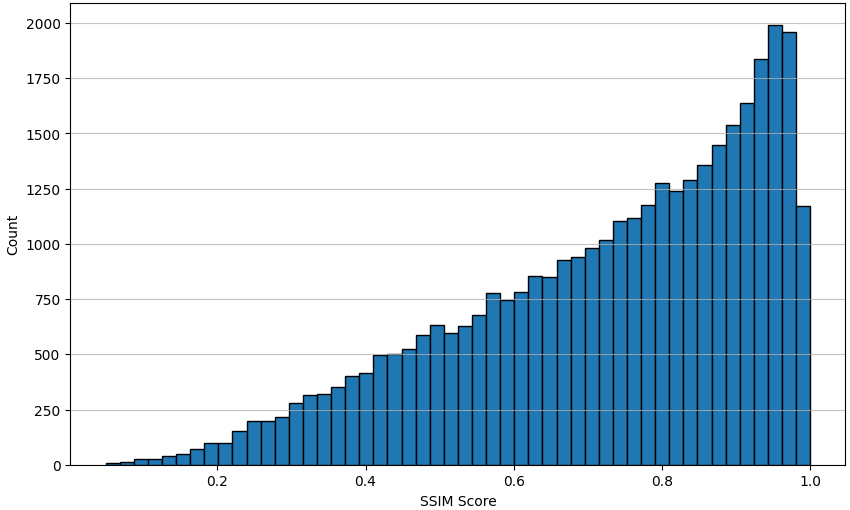}
    \caption{SSIM distribution of InstructPix2Pix training data.}
    \label{fig:ssim-ip2p}
\end{figure}

\begin{figure}[b]
    \centering
    \includegraphics[width=\columnwidth]{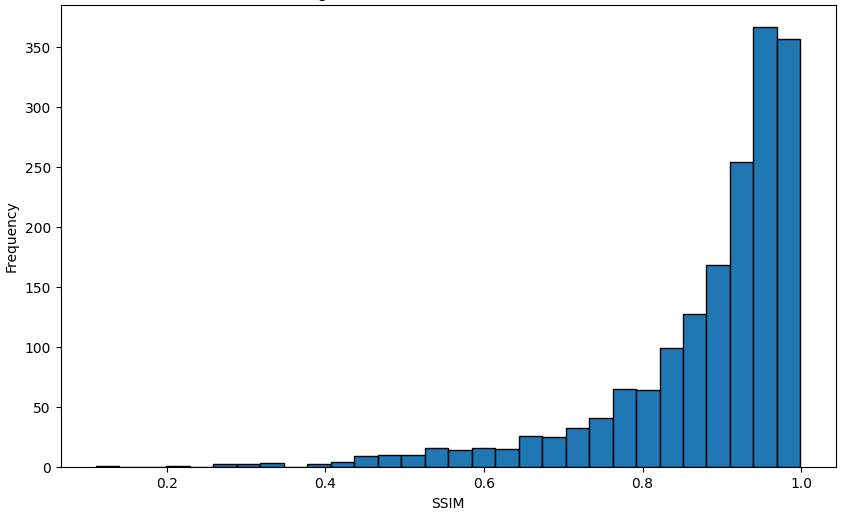}
    \caption{SSIM distribution of MagicBrush training data.}
    \label{fig:ssim-mb}
\end{figure}

\begin{figure}[b]
    \centering
    \includegraphics[width=\columnwidth]{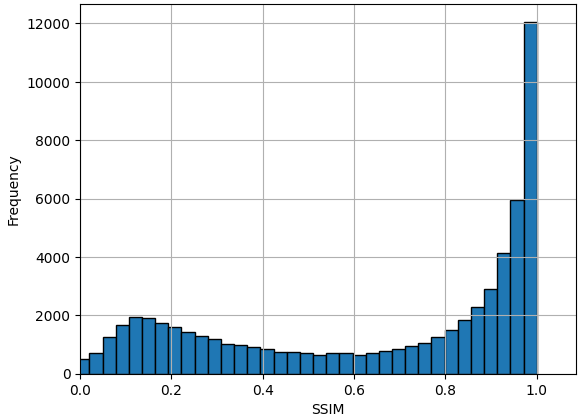}
    \caption{SSIM distribution of \ours training data.}
    \label{fig:ssim-us}
\end{figure}

We observe that human-generated edits often introduce substantial diversity, such as rearranging objects or people, which significantly impacts Structural Similarity Index Measure (SSIM) scores. These variations create a distributional mismatch with InstructPix2Pix’s pretraining data (Figures \ref{fig:ssim-ip2p}, \ref{fig:ssim-mb}, \ref{fig:ssim-us}), where edits are generally more constrained. To better understand this difference, we analyze SSIM distributions, highlighting the gap between human edits and the structured outputs of synthetic datasets.

\begin{figure*}[h]
    \centering
    \includegraphics[width=\textwidth]{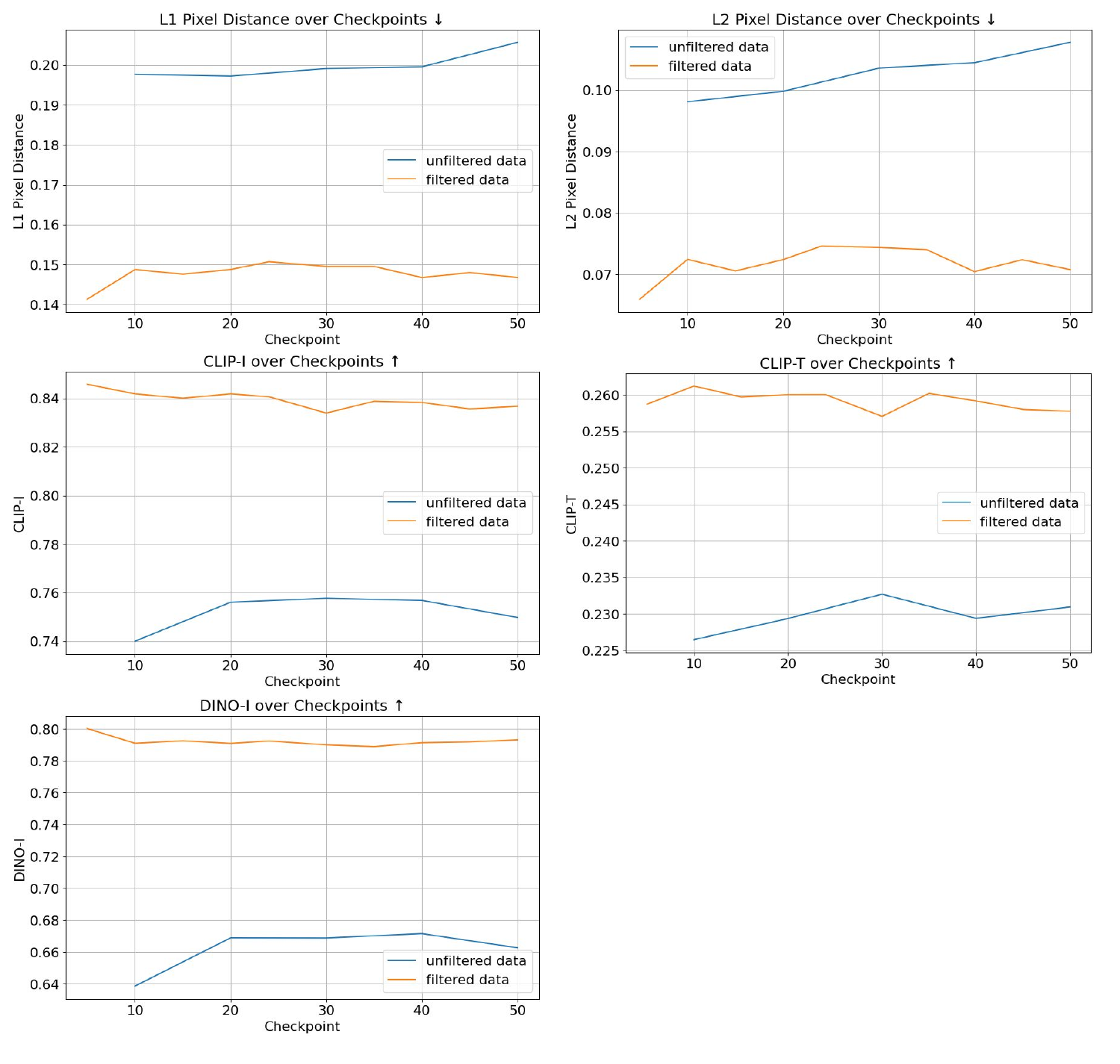}
    \caption{Filtering the data massively improved CLIP-based metrics.}
    \label{fig:filtered_data}
\end{figure*}

To make our dataset more compatible with InstructPix2Pix, we currently apply SSIM-based filtering to exclude edits that deviate too far from the pretraining distribution. Following this, we use the same CLIP-based filtering methodology employed by InstructPix2Pix to further refine the data. We verify that this filtering leads to a more capable model using the VIE-scores (Table \ref{tab:filtered-data}) and CLIP-based metrics (Figure \ref{fig:filtered_data}). Our approach relies on thresholding to identify and remove outliers, but we recognize that soft sampling techniques could offer a more flexible and nuanced alternative. Exploring such methods remains a promising direction for future work.

\input{tables/filtered_data}

\subsection{Processing instructions}
\begin{figure*}[h]
    \centering
    \includegraphics[width=\textwidth]{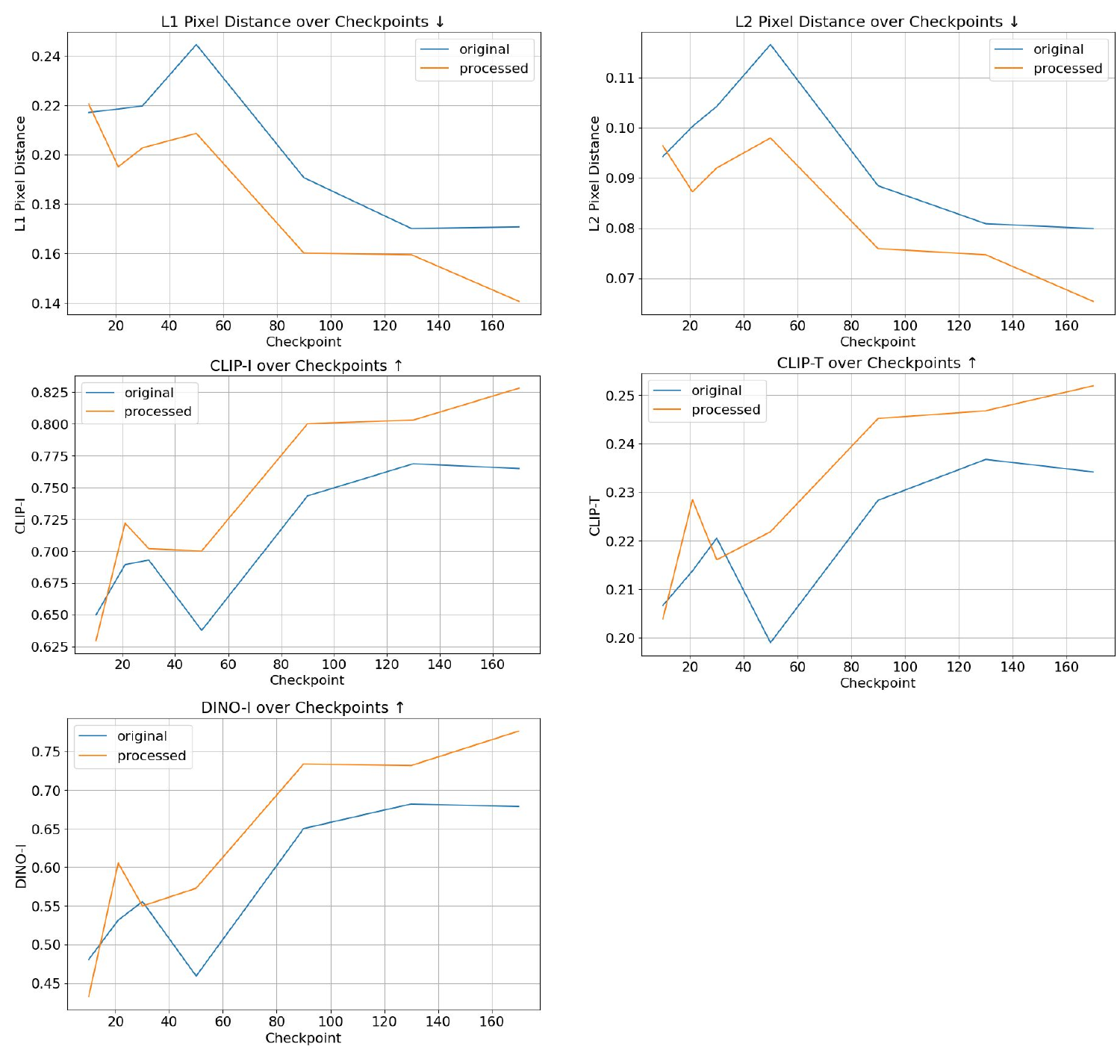}
    \caption{Processing instructions consistently yields better results on CLIP-based results.}
    \label{fig:processed_instruction}
\end{figure*}

Reddit users often provide vague, unclear instructions with unnecessary details, hindering the editing process. To address this, we refined these instructions for greater clarity and relevance. To evaluate the impact of this preprocessing, we trained two models under the same conditions: one with the original instructions and the other with the processed versions. Results in Table \ref{tab:process_insructions} and Figure \ref{fig:processed_instruction} show that these have a significant effect on model performance. 

We ran this experiment early in the development processes with a suboptimal training strategy and a smaller subset of the data, leading to much lower scores compared to our final model.

\input{tables/processed_instructions}

% \begin{flushleft}
% \newpage
% \end{flushleft}

% \clearpage

% \clearpage

\begin{figure}[h]
    \centering
    \includegraphics[width=\linewidth]{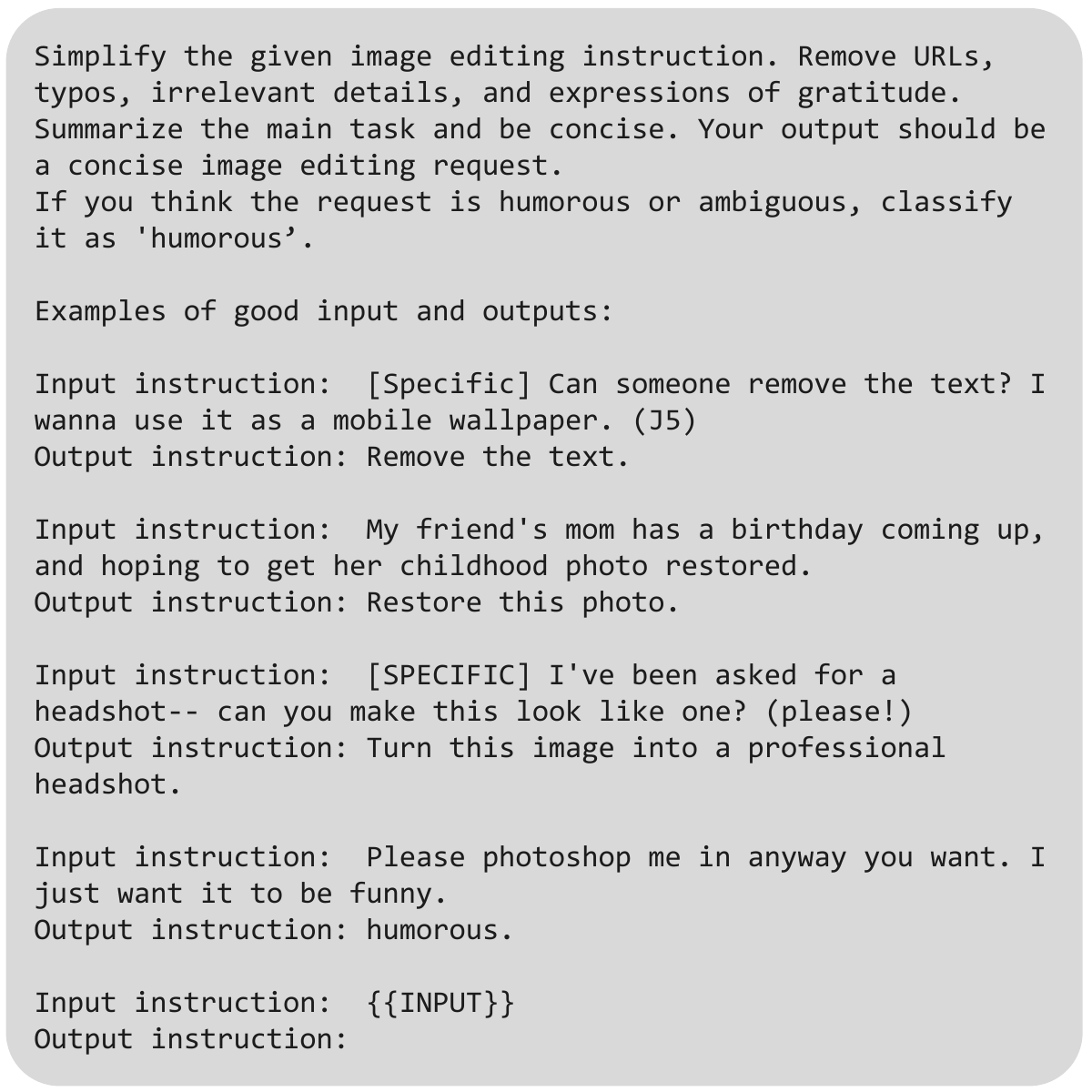}
    \caption{GPT-4o prompt for instruction rewriting.}
    \label{fig:instruction_rewriting}
\end{figure}

\section{Inference time results}
\subsection{Hyperparameters}

We conducted several inference-time experiments: varying the number of diffusion steps, the image and text guidance scales, and further rewriting instructions with GPT-4o to add more details. 

See equation (6) in ~\cite{ho2022classifier} for the definition of classifer-free guidance scale. The conventional wisdom is that higher image guidance scale make the generated image look more similar to the original image, while higher text guidance scale improve instruction adherence. Additionally, higher number of inference steps are believed to improve the quality of the generated image at the expense of computational time. Our statistical experiments do not capture these relationships, and even demonstrate the opposite relationship in case of image guidance scale.

\begin{figure}[h]
    \centering
    \includegraphics[width=\columnwidth]{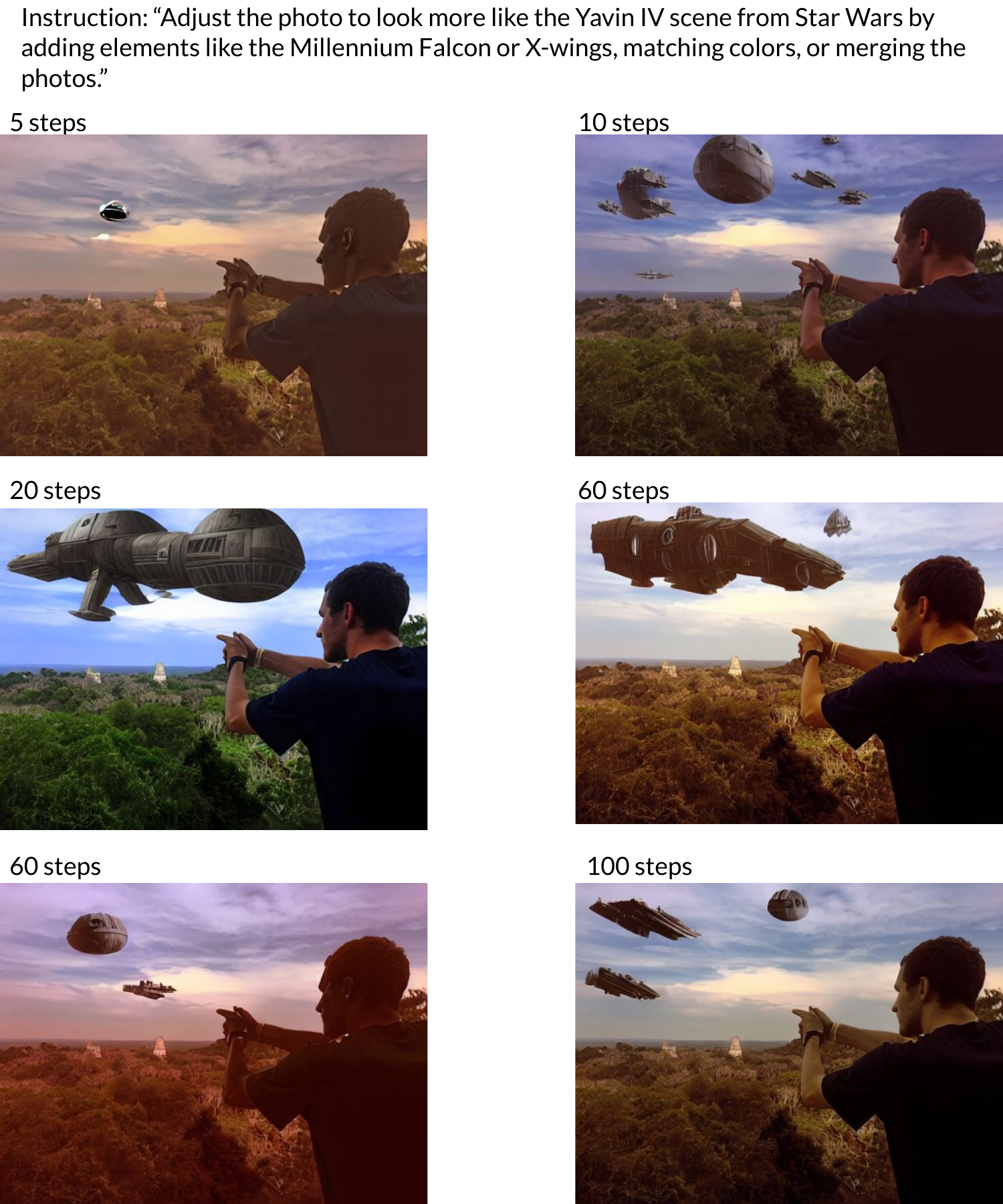}
    \caption{Increasing the number of diffusion steps above 20 usually does not improve the quality.}
    \label{fig:inference_steps_example}
\end{figure}

\begin{figure}[h]
    \centering
    \includegraphics[width=\columnwidth]{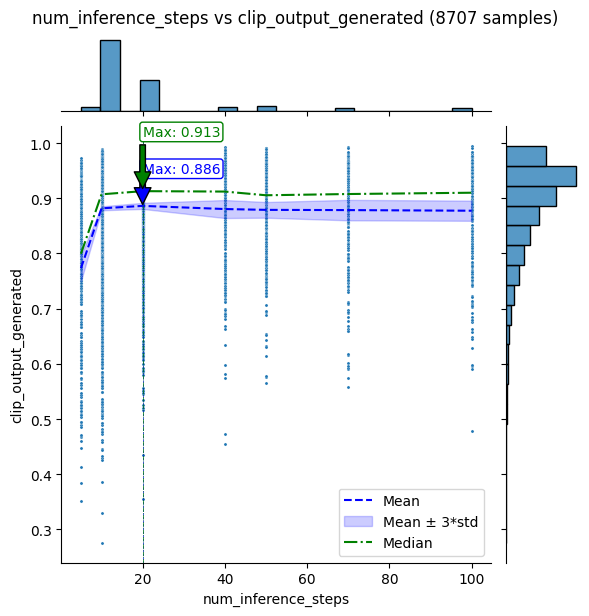}
    \caption{The number of inference steps does not improve the generated image quality, as measured by the CLIP similarity between the generated image and the most upvoted Reddit edit.}
    \label{fig:num_inference_steps}
\end{figure}

\paragraph{Number of inference steps}
We observe that 20 inference steps strike a good balance between the computational time and the image quality. Specifically, we find that the average CLIP similarity between the generated image and the most upvoted Reddit edit is approximately the same for any setting of inference steps above 20. See Figure \ref{fig:num_inference_steps} for the statistical plot and figure \ref{fig:inference_steps_example} for an example.

\begin{figure}[h]
    \centering
    \includegraphics[width=\columnwidth]{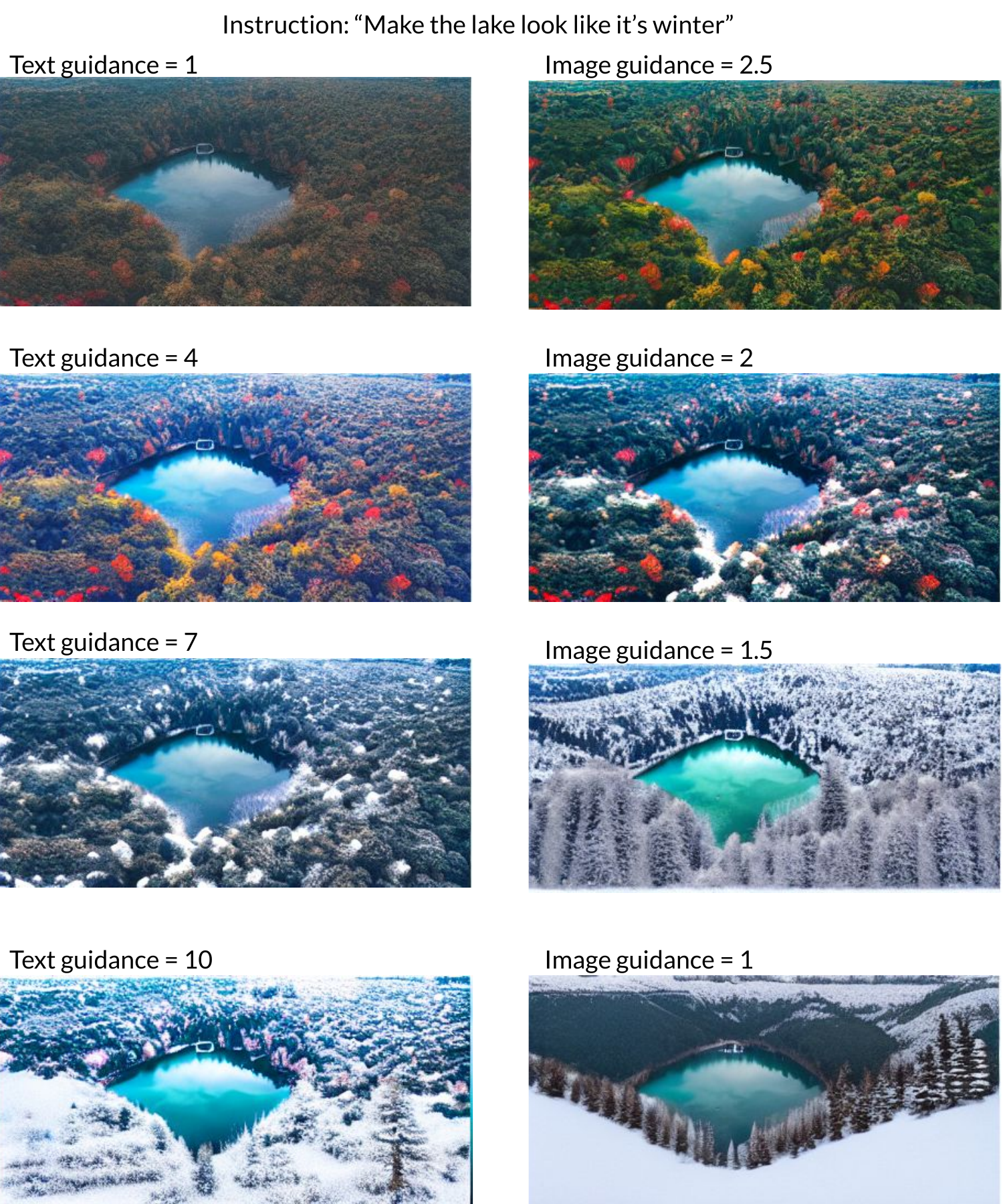}
    \caption{An example where guidance scales behave as expected.}
    \label{fig:guidance_scales_example}
\end{figure}

\begin{figure}[h]
    \centering
    \includegraphics[width=\columnwidth]{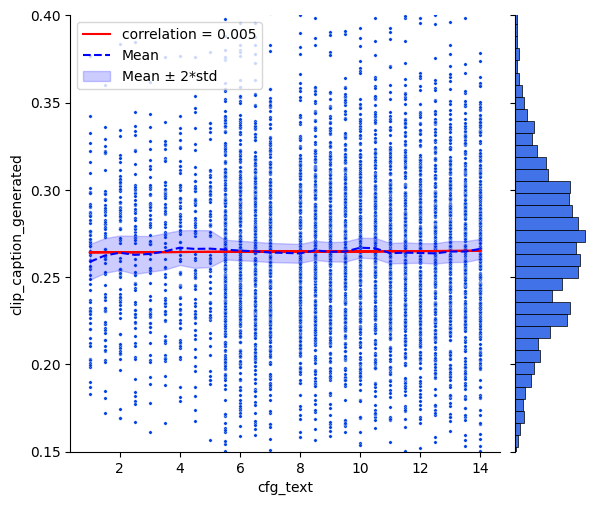}
    \caption{Text guidance scale has no effect on instruction adherence, as measured by the CLIP similarity between the generated image and the caption of the expected output, as in figure \ref{fig:test_set_figure}.}
    \label{fig:text_guidance_scale}
\end{figure}

\paragraph{Text guidance scale}
We observe \textbf{no correlation} ($\rho=.005$) between the text guidance scale in range $[1,14]$ and instruction adherence, as measured by CLIP similarity between the generated image and the caption describing the desired output. See Figure \ref{fig:text_guidance_scale}. While there is no correlation in aggregate, some individual edits may still change significantly with different text guidance scales, see Figure \ref{fig:guidance_scales_example} for such an example.

\begin{figure}[h]
    \centering
    \includegraphics[width=\columnwidth]{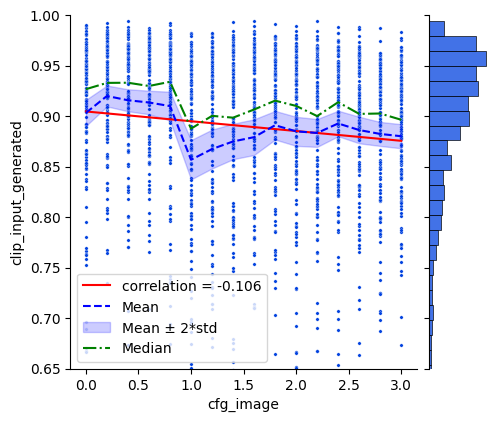}
    \caption{Increased image guidance scale results in \textbf{less} similar images, as measured by CLIP similarity between the input and generated images.}
    \label{fig:image_guidance_scale}
\end{figure}

\paragraph{Image guidance scale}
The generated image quality decreases sharply if the image guidance scale is above 3. Inside the $[1,3]$ range, the image scale makes little difference in aggregate. Counter-intuitively, we observe a \textbf{negative} correlation ($\rho=-.106$) between image guidance scale and CLIP similarity between the input and generated images. In other words, higher image guidance values result in \textbf{less similar} images on average, which contradicts conventional assumptions about guidance scales and warrants further investigation. See Figure \ref{fig:image_guidance_scale}.

\subsection{Instruction rewriting} 
As the diffusion model lacks reasoning capabilities, it often fails when asked to interpret abstract or creative instructions. To improve outcomes on these examples, we employ a large language model (LLM) to rewrite instructions in a more specific manner, similar to Dalle-3 ~\cite{betker2023dalle3}. Since only creative edit tasks benefit from this technique, we do not make this part of our main pipeline. We gave the input image and the original instruction to GPT-4o with the prompt ``\textit{You are given an image editing instruction. If the instruction is already concrete and specific, do not rewrite it at all. If the instruction is vague or does not make sense for the image, then rewrite it. Make the new instruction specific and detailed, e.g. do not use words 'enhance', 'adjust', 'any'.}"

\begin{figure*}[h]
    \centering
    \includegraphics[width=\textwidth]{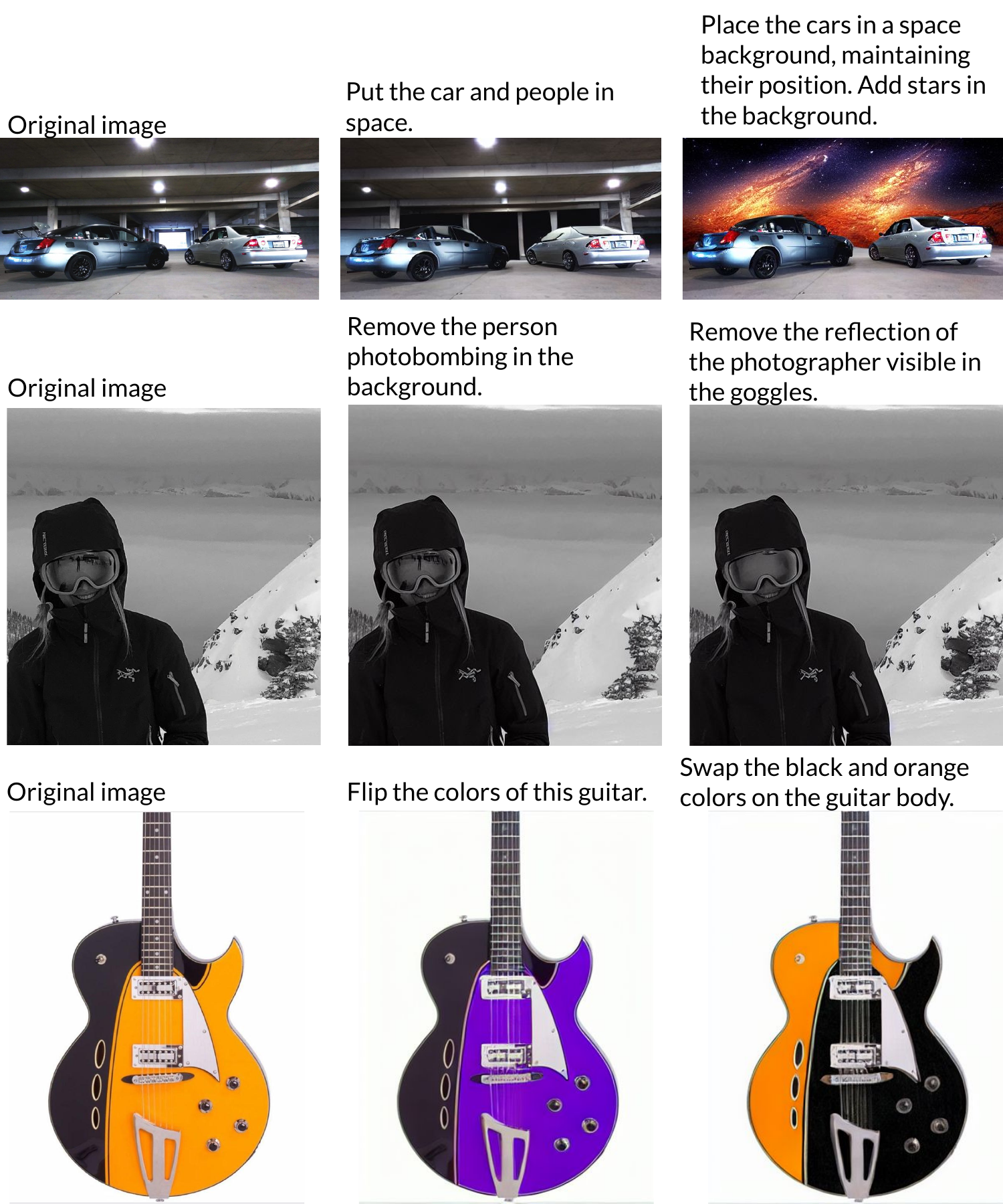}
    \caption{Detailed instructions can improve edit quality on certain classes of tasks.}
    \label{fig:instruction_rewrite}
\end{figure*}

\subsection{Quantitative evaluation on external test sets}
\label{supp:ood_eval}
\input{tables/mb_emu_metrics}

Despite being out of distribution, the \ours model performs comparably to other models on the synthetic datasets Emu Edit~\cite{sheynin2024emu} and MagicBrush~\cite{zhang2024magicbrush}. On several metrics (VQA\_CLIP and TIFA on MagicBrush and VQA\_llava, VQA\_Flan-t5 and TIFA on Emu Edit), the \ours model is within 1 standard deviation of the highest scoring model, indicating that it is fairly generalizable to new tasks. 

\subsection{Elo scores}

\begin{figure*}
    \centering
    \includegraphics[width=\textwidth]{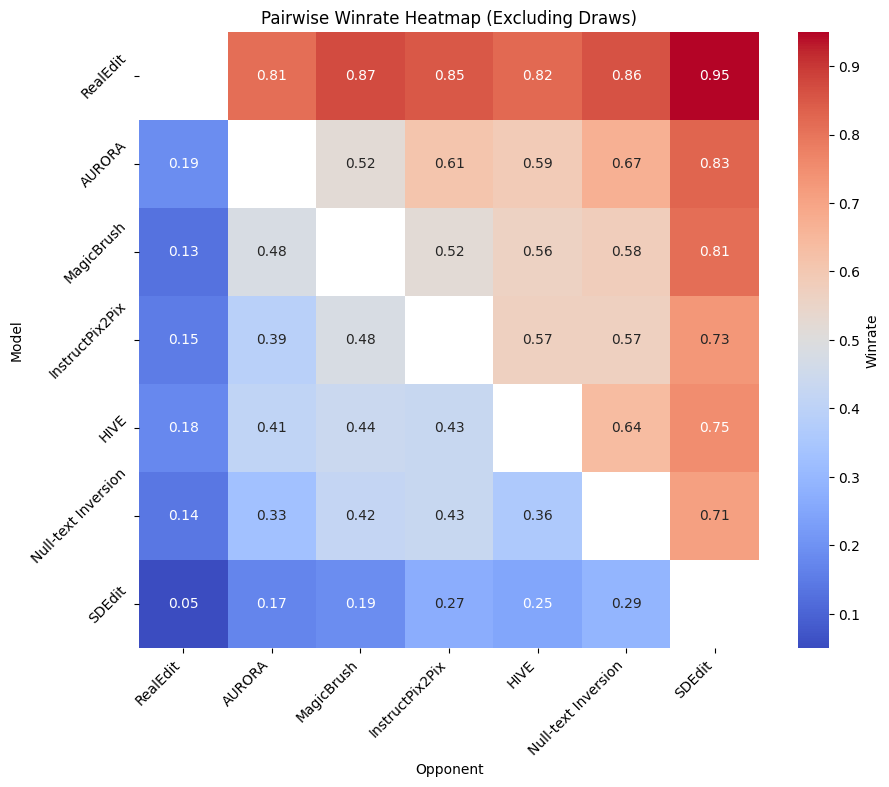}
    \caption{\textbf{Heatmap of pairwise winrates on our test set.} We excluded draws for this heatmap. }
    \label{fig:elo_heatmap}
\end{figure*}

To evaluate Elo scores, we leverage Amazon Mechanical Turk (MTurk) for conducting pairwise comparisons. We selected 200 diverse examples from our dataset to ensure coverage of various editing tasks and performed comparisons across all seven models in our benchmark. This process resulted in a total of 4,200 pairwise evaluations, providing a robust dataset for assessing human preferences. We present a table of pairwise winrates (Figure \ref{fig:elo_heatmap})  

In addition to evaluating our dataset, we extended our analysis to the Imagen Hub Museum\cite{ku2024imagenhub} tasks, building on the results from the GenAI Arena\cite{jiang2024genai}. Using their generations, available on HuggingFace, we incorporated results from our model to facilitate direct comparisons. For these evaluations, we conducted a new round of pairwise comparisons where we matched one model from their benchmark against our model for the same tasks. This allowed us to directly assess how our model performs relative to state-of-the-art models on external datasets.

The evaluations on MTurk followed a structured protocol to ensure reliability and consistency. Workers were asked to compare image outputs based on task completion, realism, and alignment with instructions. The use of MTurk enabled us to gather diverse human feedback efficiently and at scale. The full results are presented in Table \ref{tab:elo_genai}, highlighting the comparative performance across different models.

\input{tables/elo_genai}

% \clearpage

\begin{figure*}
    \centering
    \includegraphics[width=\textwidth]{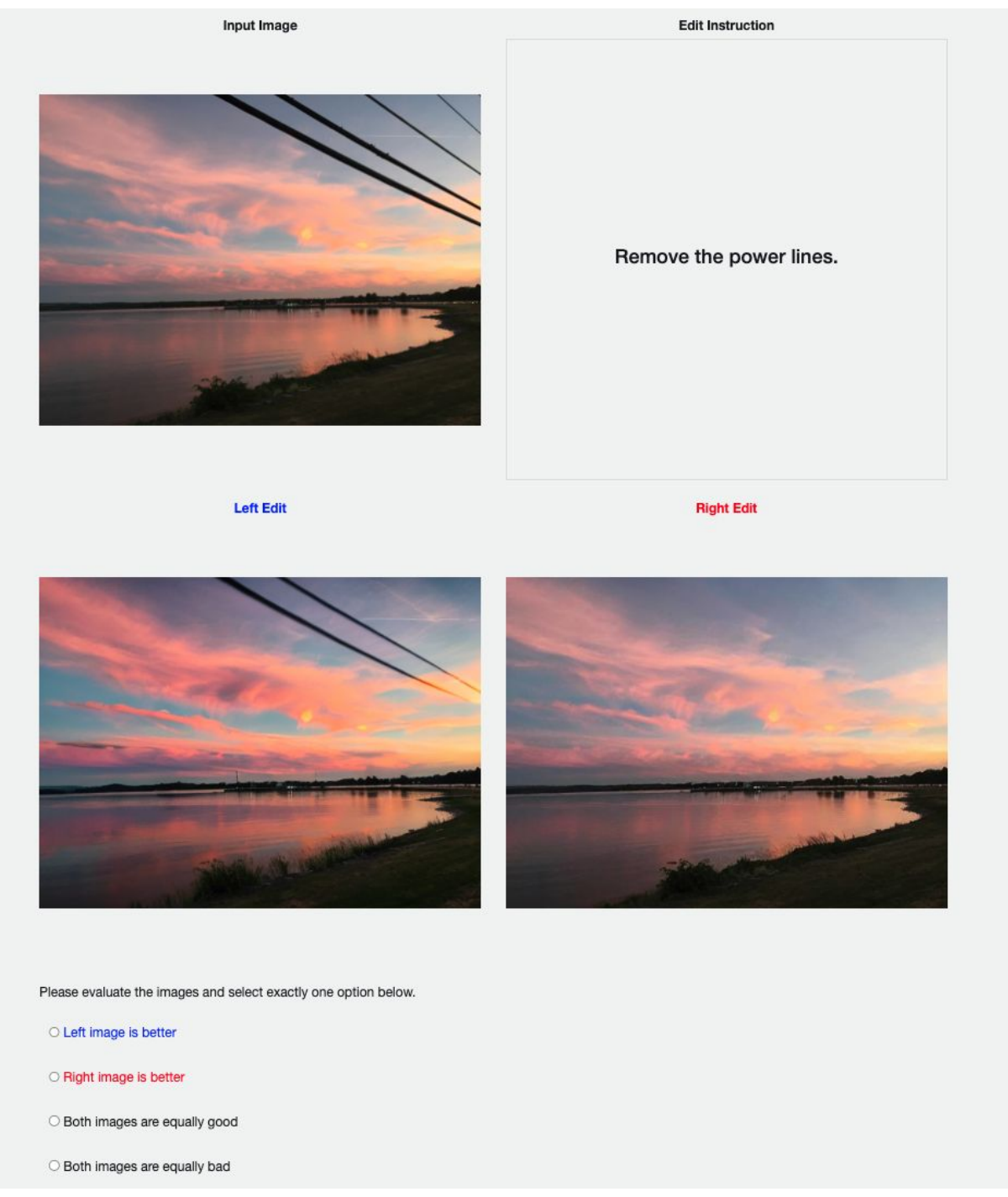}
    \caption{\textbf{Interface for Elo evaluation on MTurk}. To complete Elo evaluations, we hired workers on Amazon Mechanical Turk to compare the quality of different editing models.}
    \label{fig:mturk_interface}
\end{figure*}

% \clearpage

\section{Reddit experiment}

To evaluate the generalization capability of our model, we deployed it on Reddit. Specifically, we targeted two subreddits: r/PhotoshopRequest and r/estoration, which focus on image editing and restoration tasks. Adhering to the community guidelines of these subreddits, we collected posts requesting image edits and processed them using our model.

\begin{figure}[h]
    \centering
    \includegraphics[width=\columnwidth]{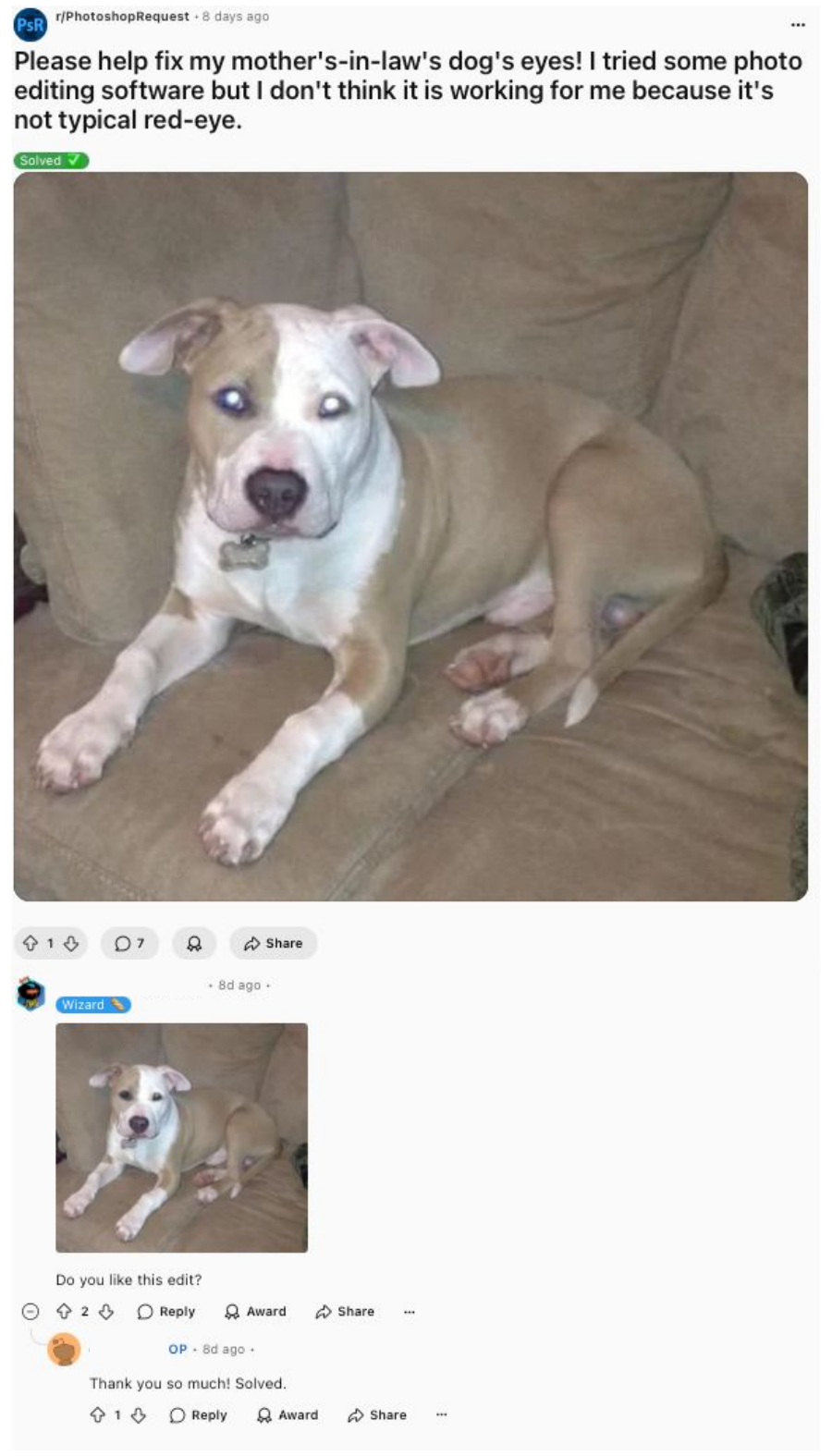}
    \caption{\textbf{Our model successfully completes new requests on Reddit.} Deployed on the original subreddits, it handled in-the-wild requests effectively as seen by OP's response.}
    \label{fig:reddit_dog}
\end{figure}

\begin{figure}[h]
    \centering
    \includegraphics[width=\columnwidth]{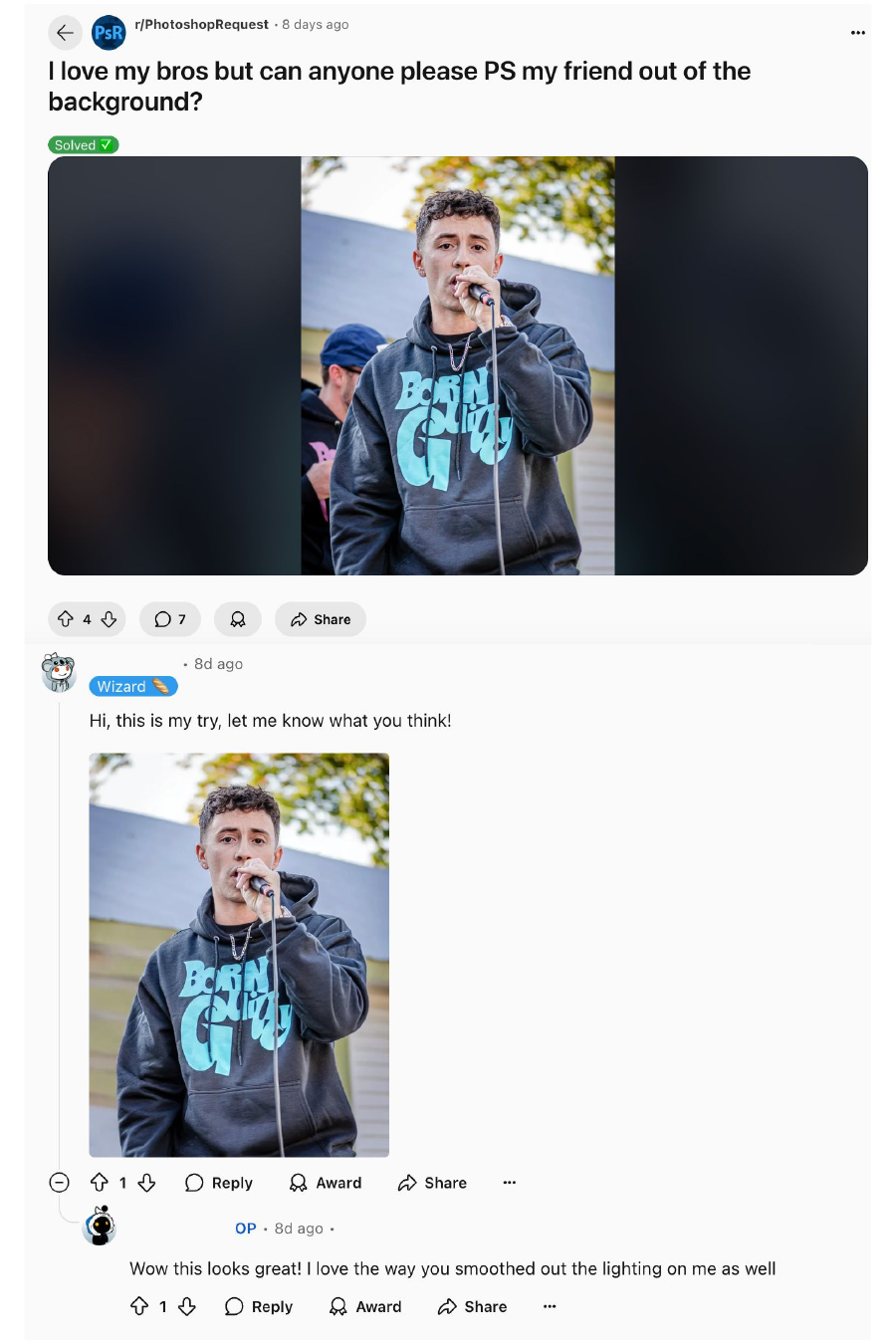}
    \caption{\textbf{Our model successfully completes new requests on Reddit.} Deployed on the original subreddits, it handled in-the-wild requests effectively as seen by OP's response.}
    \label{fig:reddit_man}
\end{figure}

For each processed request, we submitted a comment containing the generated output image along with a brief message asking for feedback from the original poster (OP). With this experiment, we gathered qualitative evaluations from humans, and provide insight into the model's performance in real world scenarios. See Figures \ref{fig:reddit_dog}
and \ref{fig:reddit_man}.  

\clearpage
\section{Edited image detection}

% \subsection{Implementation details}

% \noindent\textbf{Model architecture}

% Universal Fake Detect (UFD) processes images into latent representations using CLIP and performs binary classification through linear probing. The original model by ~\citet{ojha2024universalfakeimagedetectors} employs a single linear layer, whereas \truemedia enhances the architecture by adding a second linear layer with a ReLU activation in between. \truemedia continues to utilize the CLIP:ViT-L/14 encoder with a resolution of 224x224. The input dimension of the first linear layer is 768, and the second linear layer has a dimension of 384. The classifier predicts an image as fake (positive class) when the output logit exceeds a threshold of 0.5.

% Universal Fake Detect (UFD) works by processing images into latent representations via CLIP and then applying linear probing for binary classification. The original model trained by ~\citet{ojha2024universalfakeimagedetectors} uses one linear layer while \truemedia adds a second linear layer with a ReLU in between;\truemedia also continues using the 224 CLIP:ViT-L/14 encoder. The input dimension for the first linear layer is 768 and the second layer has dimension 384. The model makes a fake (positive class) prediction when the single output logit exceeds the threshold of 0.5. \\

% This is the only architectural difference. 

\paragraph{Data processing and training}
\label{UFD_training}
The baseline classifier undergoes a multi-stage training process: initially on academic datasets and subsequently fine-tuned on \truemedia's proprietary data. In total, the baseline model is trained on 65K images with a near equal 50/50 split between real and generated images. To assess the value of \ours data for fake image detection, we train a second version of UFD by combining the original data with \ours data. Specifically, we include only photographs, excluding non-photographic images such as digital artworks, screenshots, cartoons, and infographics, filtered using GPT-4o. This single-stage training incorporates an additional 37K original and 37K edited images, resulting in a total of 139K images.

In the first stage of training, the \truemedia model took over 24 hours to train on an A10G GPU with 20GB of RAM and the remaining three stages took 4 hours. Our optimized model took 1.5 hours to train on a L40S GPU with 40GB of RAM.\\

\begin{table}[h]
    \centering
    \caption{Breakdown of fake image sources in the training recipe of the \truemedia model used as our baseline.}
    \label{tab:fake-sources}
    \resizebox{\linewidth}{!}{%
    \begin{tabular}{l|c}
        \toprule
        \textbf{Source} & \textbf{Count} \\ \midrule
        DiffusionDB ~\cite{wang2023diffusiondblargescalepromptgallery} & 16K \\ 
        StyleGAN2-FFHQ ~\cite{karras2020analyzingimprovingimagequality} & 8K \\ 
        Stable-Diffusion-Face ~\cite{sdfd} (512 resolution) & 2.4K \\ 
        Stable-Diffusion-Face (768 resolution) & 2.4K \\ 
        Stable-Diffusion-Face (1024 resolution) & 2.4K \\ 
        Fakes uploaded to \truemedia & 2K \\ \bottomrule
        % (2/20–8/15)
    \end{tabular}%
    }
\end{table}

\begin{table}[h]
    \centering
    \caption{Breakdown of real image sources in the training recipe of the baseline model.}
    \label{tab:real-sources}
    \resizebox{\linewidth}{!}{%
        \begin{tabular}{l|c}
        \toprule
        \textbf{Source} & \textbf{Count} \\ \midrule
        CelebA-HQ (Reals) ~\cite{karras2018progressivegrowinggansimproved} & 23K \\ 
        Random sample of COCO-Train-2017 ~\cite{lin2015microsoftcococommonobjects} & 5K \\ 
        Flickr-Faces-HQ Dataset (FFHQ) ~\cite{karras2019style} & 3K \\ 
        Reals uploaded to \truemedia  & 0.7K\\ \bottomrule
        % (2/20–8/15)
        \end{tabular}%
        }
\end{table}

% The hyperparameters for the baseline model are: batch size of 512, learning rate of 0.0001, 10 epochs, Adam optimizer with momentum ($\beta$) of 0.9, and a loss weight of 100. Data augmentation for the baseline included horizontal flipping and Gaussian noise with a $(0.0, 3.0)$ signal at a probability of 0.5. When fine-tuning on \RealEdit data, a batch size of 2048 was used, with data augmentations disabled, while the other hyperparameters remained unchanged.

% The baseline model uses the following hyperparameters: a batch size of 512, a learning rate of 0.0001, 10 epochs, the Adam optimizer with a momentum ($\beta$) of 0.9, and a loss weight of 100. Data augmentation includes horizontal flipping and Gaussian noise with a $(0.0, 3.0)$ signal, applied with a probability of 0.5. For fine-tuning on \RealEdit data, the batch size is increased to 2048, and data augmentations are disabled, while all other hyperparameters remain unchanged.

% The hyper-parameters for the baseline model are: batch size of 512, learning rate of 0.0001, 10 epochs, Adam optimizer with momentum ($\beta$) of 0.9, and a loss weight of 100.

% For the baseline model, data augmentation included horizontal flipping and Gaussian noise with a $(0.0, 3.0)$ signal at a probability of 0.5. When fine-tuning the baseline on \RealEdit data, a batch size of 2048 was used, with data augmentations disabled. The remaining hyperparameters were kept unchanged.

% \noindent\textbf{Hardware}

% \noindent\textbf{\truemedia's In-the-wild test set}

\paragraph{\truemedia's in-the-wild test set}

\truemedia's in-the-wild test set includes images uploaded between 8/16/2024 and 11/10/2024. We randomly sample 100  real images and then sample 100 fake images selected from those tagged as "likely photoshopped" by professional sleuths in \truemedia's media database, ensuring the evaluation focuses on human-edited images rather than exclusively AI-generated edits. Tool usage data was available for some images, revealing that approximately 80\% of the fake images were human edits created with Photoshop, while the remaining 20\% involved human edits combined with AI tools such as Dream Studio AI, Insightface AI, and Remaker AI.

\paragraph{Qualitative example}

\begin{figure}[t]
    \centering
    \includegraphics[width=\columnwidth]{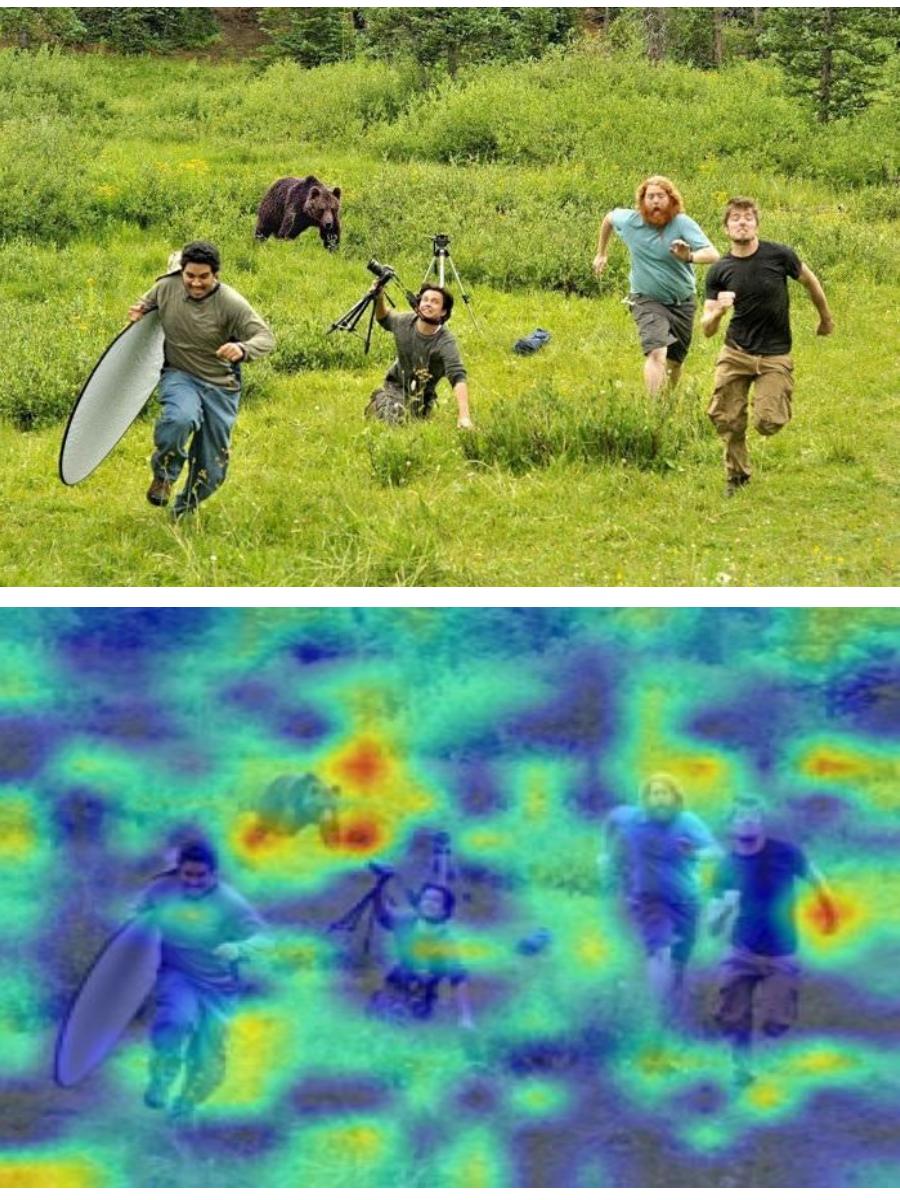}
    \caption{Top: An edited image that inserted a bear to make it seem the camera crew was being chased. Bottom: Grad-CAM heat-map visualization highlighting the regions of attention.}
    \label{fig:bear_camera_crew}
\end{figure}

To understand how the classifier operates, we use Gradient-weighted Class Activation Mapping (Grad-CAM) \cite{Selvaraju_2019} to analyze an example. In Figure~\ref{fig:bear_camera_crew}, we show an edited image where a bear was added to the background using Photoshop. The original image did not include the bear. The baseline model incorrectly classified this photo as unedited, whereas the classifier trained with \ours data correctly identified it as edited. Grad-CAM highlights the areas of the image most influential to the classifier’s decision, as seen in the figure, where the focus is on the region around the bear. The specific implementation we adapted is from \citet{jacobgilpytorchcam}.

\onecolumn
\section{Additional results}

\begin{figure}[!h]
    \centering
    \includegraphics[width=\textwidth, height=0.8\textheight, keepaspectratio]{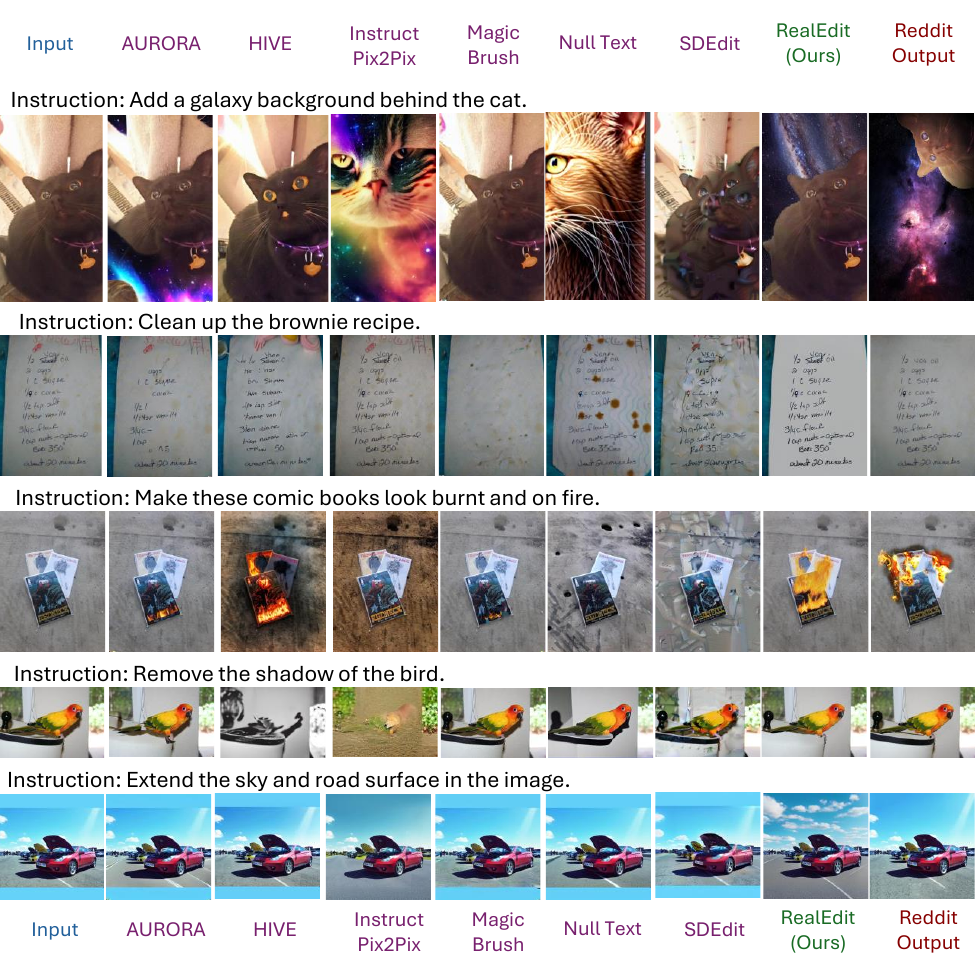}
    \caption{\textbf{Additional examples of \ours generations on \ours test set} compared to all other baseline models. We notice that the \ours model consistently outperforms other models in task completion as well as aesthetic quality.}
    \label{fig:generation_examples_2}
\end{figure}

\begin{figure*}[!htbp]
    \centering
    \includegraphics[width=\textwidth]{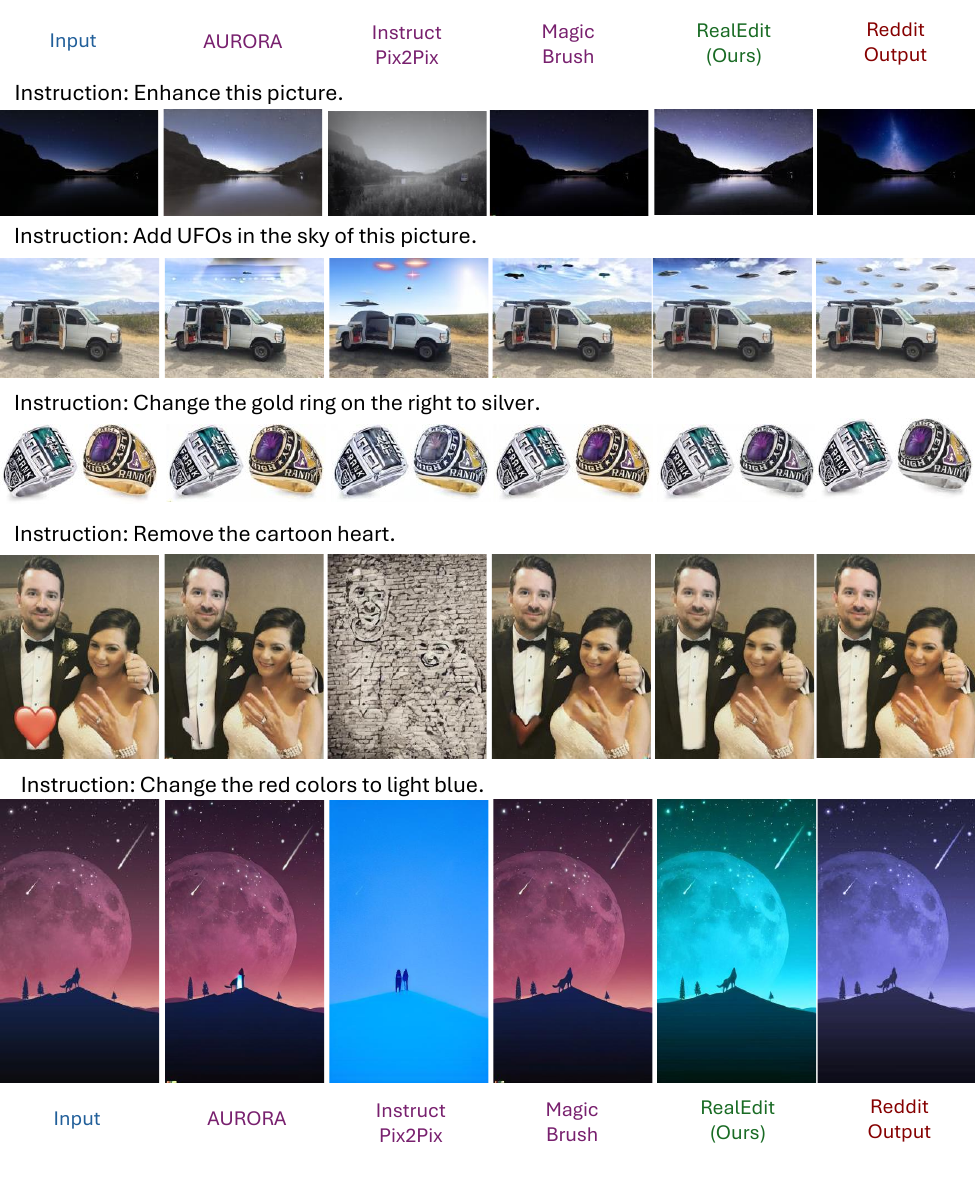}
    \caption{\textbf{Additional examples of \ours generations on \ours test set} compared to select high performing baseline models. We notice that the \ours model consistently outperforms other models in task completion as well as aesthetic quality.}
    \label{fig:generation_examples_3}
\end{figure*}
\twocolumn
% \end{document

%% file: tables/vie_by_taxonomy.tex
\begin{table*}[t]
    \centering
    \caption{\textbf{Breakdown of model performance by operation.} We find that our model is consistently best across all operations in VIE\_O, and our strongest operation is ``remove''. We use a sample of 2000 data points and take arithmetic mean of all individual scores on each data point.}
    \label{tab:vie_by_taxonomy}
    \resizebox{\textwidth}{!}{%
    \begin{tabular}{l|ccc|ccc|ccc|ccc}
    \toprule
         \multirow{2}{*}{\textbf{Operation}} & \multicolumn{3}{c|}{\textbf{AURORA}} & \multicolumn{3}{c|}{\textbf{InstructPix2Pix}} & \multicolumn{3}{c|}{\textbf{MagicBrush}} & \multicolumn{3}{c}{\textbf{RealEdit}} \\
          & \textbf{VIE\_SC} & \textbf{VIE\_PQ} & \textbf{VIE\_O}  & \textbf{VIE\_SC} & \textbf{VIE\_PQ} & \textbf{VIE\_O} & \textbf{VIE\_SC} & \textbf{VIE\_PQ} & \textbf{VIE\_O} & \textbf{VIE\_SC} & \textbf{VIE\_PQ} & \textbf{VIE\_O}\\
        \midrule
        Add         & 2.89 & 3.45 & 2.34 & 2.48 & 3.60 & 2.15 & 1.94 & \textbf{4.43} & 1.79 & \textbf{4.24} & 3.26 & \textbf{3.15} \\
        Change color & 2.38 & 3.77 & 2.26 & 2.90 & 3.61 & 2.57 & 1.95 & \textbf{4.05} & 1.83 & \textbf{5.36} & \textbf{4.05} & \textbf{4.11} \\
        Enhance     & 1.86 & 3.00 & 1.88 & 1.80 & 2.91 & 1.79 & 2.41 & 3.44 & 2.33 & \textbf{4.73} & \textbf{4.03} & \textbf{3.95} \\
        Formatting  & 0.89 & 3.02 & 0.99 & \textbf{1.70} & 3.13 & 1.31 & 0.74 & 3.57 & 0.94 & 1.66 & \textbf{4.47} & \textbf{1.66} \\
        Open ended  & 2.51 & 2.70 & 1.98 & 2.49 & \textbf{3.57} & 2.05 & 2.36 & 3.49 & 1.99 & \textbf{4.67} & 2.93 & \textbf{3.15} \\
        Remove      & 2.94 & 4.25 & 2.76 & 1.01 & 3.03 & 1.06 & 2.30 & 4.71 & 2.30 & \textbf{5.29} & \textbf{5.01} & \textbf{4.35} \\
        Replace     & 2.18 & 3.32 & 1.87 & 2.25 & 3.16 & 1.74 & 1.57 & \textbf{3.81} & 1.45 & \textbf{3.50} & 3.50 & \textbf{2.53} \\
        Restore     & 1.52 & 2.23 & 1.57 & 1.66 & 2.49 & 1.74 & 1.59 & 2.60 & 1.74 & \textbf{4.01} & \textbf{2.98} & \textbf{3.21} \\
        \bottomrule
    \end{tabular}%
    }
    
\end{table*}

%% file: tables/consistency_decoder_metrics.tex
\begin{table*}[t]
    \centering
    \caption{The decoder has minor effects on quantitative metrics but sometimes improves qualitative results.}
    \label{tab:consistency_decoder_combined}
    \resizebox{\linewidth}{!}{%
    \begin{tabular}{lccccccccc}
    \toprule
    \textbf{Model} & \textbf{VIE\_O} & \textbf{VIE\_PQ} & \textbf{VIE\_SC} & \textbf{L1} & \textbf{L2} & \textbf{CLIP-I} & \textbf{DINO-I} & \textbf{CLIP-T} \\
    \midrule
    \ours w/ original decoder & 3.54 & 3.91 & 4.37 & 0.154 & 0.069 & 0.830 & 0.782 & 0.258 \\
    \ours w/ consistency decoder & 3.48 & 3.78 & 4.34 & 0.156 & 0.069 & 0.830 & 0.779 & 0.258 \\
    \midrule
    \textbf{Change} & -0.06 & -0.13 & -0.03 & 0.002 & 0 & 0 & -0.003 & 0 \\
    \toprule
    MagicBrush w/ original decoder & 1.92 & 3.98 & 1.89 & 0.139 & 0.066 & 0.830 & 0.782 & 0.251 \\
    MagicBrush w/ consistency decoder & 1.84 & 3.93 & 1.83 & 0.135 & 0.066 & 0.831 & 0.784 & 0.251 \\
    \midrule
    \textbf{Change} & -0.08 & -0.05 & -0.06 & -0.004 & 0 & 0.001 & 0.002 & 0 \\
    \toprule
    InstructPix2Pix w/ original decoder & 1.73 & 3.37 & 1.85 & 0.183 & 0.075 & 0.754 & 0.651 & 0.243 \\
    InstructPix2Pix w/ consistency decoder & 1.89 & 3.40 & 1.95 & 0.180 & 0.073 & 0.758 & 0.648 & 0.244 \\
    \midrule
    \textbf{Change} & 0.16 & 0.03 & 0.10 & -0.003 & -0.002 & 0.004 & -0.003 & 0.001 \\
    \bottomrule
    \end{tabular}%
    }
\end{table*}

%% file: tables/filtered_data.tex
\begin{table}[!ht]
    \centering
    \caption{Aligning \ours data to the pretraining distribution yields better results.}
    \label{tab:filtered-data}
    \begin{tabular}{lccc}
    \toprule
    \textbf{Model} & \textbf{VIE\_O} & \textbf{VIE\_PQ} & \textbf{VIE\_SC} \\
    \midrule
    Filtered data & \textbf{3.48} & \textbf{3.78} & \textbf{4.34} \\
    Original data & 2.35 & 2.99 & 2.91 \\
    \bottomrule
    \end{tabular}
\end{table}

%% file: tables/processed_instructions.tex
\begin{table}[!h]
    \centering
    \caption{Processing instructions improves model performance.}
    \label{tab:process_insructions}
    \begin{tabular}{lccc}
    \toprule
    \textbf{Model} & \textbf{VIE\_O} & \textbf{VIE\_PQ} & \textbf{VIE\_SC} \\
    \midrule
    Processed instructions & \textbf{2.42} & \textbf{3.72} & \textbf{2.84} \\
    Original instructons & 2.06 & 3.10 & 2.45 \\
    \bottomrule
    \end{tabular}
\end{table}

%% file: tables/mb_emu_metrics.tex
\begin{table*}[b]
    \centering
    
    \caption{\textbf{Evaluation on MagicBrush and Emu Edit test sets.} All scores within 1 standard deviation of the highest score are underlined. The \RealEdit model is still able to perform competitively on some metrics despite these tasks being out of distribution.}
    \label{tab:mb_emu_benchmark}
    \resizebox{\textwidth}{!}{%
    \begin{tabular}{lcccccc|cccccc}
    \toprule
    & \multicolumn{6}{c|}{\textbf{MagicBrush Test Set}} & \multicolumn{6}{c}{\textbf{Emu Edit Test Set}}\\
    \midrule
    \textbf{Model}  & \textbf{VIES\_SC $\uparrow$} & \textbf{VIEPQ $\uparrow$} & \textbf{VIEO $\uparrow$} & \textbf{VQA\_llava $\uparrow$} & \textbf{VQA\_CLIP $\uparrow$} & \textbf{TIFA $\uparrow$} & \textbf{VIES\_SC $\uparrow$} & \textbf{VIE\_PQ $\uparrow$} & \textbf{VIE\_O $\uparrow$} & \textbf{VQA\_llava $\uparrow$} & \textbf{VQA\_Flan-t5 $\uparrow$} & \textbf{TIFA $\uparrow$} \\
    \midrule
    AURORA~\cite{krojer2024learning}            &  \textbf{4.11} & 3.86 & \textbf{5.52} & 0.5179 & \underline{0.6517} & \underline{0.6968} & 3.40 & \underline{4.86} & 3.81 & \underline{0.4923} & \underline{0.6178} & \textbf{0.6705}\\
    Emu Edit~\cite{sheynin2024emu}     &   N/A & N/A & N/A & N/A & N/A & N/A & \textbf{4.66} & \underline{5.11} & \textbf{5.72} & \textbf{0.5130} & \textbf{0.6489} & \underline{0.6692}\\
    HIVE~\cite{zhang2024hive}                   &  2.86 & \textbf{5.02} & 3.43  & \underline{0.5200} & \underline{0.6547} & \underline{0.6918} & 1.89 & \textbf{5.50} & 2.06 & 0.4372 & 0.5258 & 0.6447\\
    InstructPix2Pix~\cite{brooks2023instructpix2pix}  & 2.63 & \underline{4.70} & 3.06 & 0.4490 & 0.5518 & 0.6615 & 2.15 & \underline{5.00} & 2.36 & 0.4261 & 0.5061 & 0.6343\\
    MagicBrush~\cite{zhang2024magicbrush}       & \underline{3.43} & \underline{4.89} & 4.11 & \textbf{0.5554} & \textbf{0.7138} & \textbf{0.7103} & 2.91 & \underline{5.47} & 3.13 & 0.4680 & 0.5808 & \underline{0.6628} \\
    Null-text Inv.~\cite{mokady2023null}    & 2.77 & \underline{4.74} & 3.29 & \underline{0.5246} & \underline{0.6429} & \underline{0.6899} & 3.43 & \underline{5.10} & 3.93 & 0.4823 & 0.5931 & \underline{0.6578}\\
    SDEdit~\cite{meng2021sdedit}               & 0.90 & 2.26 & 1.02 & 0.4185 & 0.4191 & 0.6167 & 0.95 & 3.23 & 1.06 & 0.4406 & 0.5145 & 0.6417\\
    RealEdit                                  & 3.12 & 3.60 & 4.09 & 0.5088 & \underline{0.6299} & \underline{0.6865} & 3.27 & \underline{4.86} & 3.84 & \underline{0.4938} & \underline{0.6158} & \underline{0.6650}\\ 

    \bottomrule
    % \multicolumn{7}{c}{\textbf{Emu Edit Test Set}}\\
    % \midrule
    % \textbf{Model}         & \textbf{VIES\_SC} $\uparrow$ & \textbf{VIE\_PQ} $\uparrow$ & \textbf{VIE\_O} $\uparrow$ & \textbf{VQA\_llava} $\uparrow$ & \textbf{VQA\_CLIP} $\uparrow$ & \textbf{TIFA} $\uparrow$\\
    % \midrule
    % AURORA~\cite{krojer2024learning}       & 3.40 & 4.86 & 3.81 & 0 & 0 & 0\\
    % HIVE~\cite{zhang2024hive}         & 1.89 & 5.5 & 2.06 & 0 & 0 & 0\\
    % InstructPix2Pix~\cite{brooks2023instructpix2pix}  & 2.15 & 5.0 & 2.36 & 0 & 0 & 0\\
    % MagicBrush~\cite{zhang2024magicbrush}       & 2.91 & \textbf{5.47} & 3.13 & 0 & 0 & 0 \\
    % Null-text Inv.~\cite{mokady2023null}  & 0 & 0 & 0 & 0 & 0 & 0\\
    
    % SDEdit~\cite{meng2021sdedit}       & 0.95 & 3.23 & 1.06 & 0 & 0 & 0\\

    % Emu-Edit~\cite{sheynin2024emu}        & \textbf{4.66} & 5.11 & \textbf{5.72} & 0 & 0 & 0\\
    % RealEdit        & 3.27 & 4.86 & 3.84 & 0 & 0 & 0\\
    % \bottomrule
    \end{tabular}%
    }
\end{table*}

%% file: tables/elo_genai.tex
\begin{table}[ht]
    \centering
    \caption{Elo scores of models based on the GenAI\cite{jiang2024genai} test set.}
    \begin{tabular}{lccc}
        \toprule
        Model & Elo Rating & 95\% C.I. & Sample Size \\
        \midrule
        MagicBrush~\cite{zhang2024magicbrush}      & 1107 & -39/+47  & 132 \\
        CosXLEdit~\cite{stabilityai2024cosxl}       & 1064 & -49/+42  & 133 \\
        RealEdit        & 1043 & -12/+17  & 1117 \\
        InfEdit~\cite{xu2023infedit}         & 1023 & -44/+39  & 122 \\
        InstructPix2Pix~\cite{brooks2023instructpix2pix} & 1011 & -50/+47  & 117 \\
        Prompt2prompt~\cite{hertz2022prompt}   & 1011 & -46/+46  & 119 \\
        PNP~\cite{tumanyan2023plug}             & 992  & -43/+62  & 122 \\
        SDEdit~\cite{meng2021sdedit}          & 991  & -48/+35  & 126 \\
        CycleDiffusion~\cite{wu2023latent}  & 933  & -41/+49  & 120 \\
        Pix2PixZero~\cite{parmar2023zero}     & 834  & -46/+41  & 126 \\
        \bottomrule
    \end{tabular}
    \label{tab:elo_genai}
\end{table}